\def\year{2019}\relax
\documentclass[letterpaper]{article} 
\usepackage{aaai19}  
\usepackage{times}  
\usepackage{helvet}  
\usepackage{courier}  
\usepackage{url}  
\usepackage{graphicx}  
\usepackage{array}
\usepackage{amssymb,amsmath}
\usepackage{ctable,multirow,makecell}
\newcommand*\rot{\rotatebox{90}}
\usepackage{xspace}
\makeatletter
\DeclareRobustCommand\onedot{\futurelet\@let@token\@onedot}
\def\@onedot{\ifx\@let@token.\else.\null\fi\xspace}
\def\eg{\emph{e.g}\onedot} 
\def\ie{\emph{i.e}\onedot}

\def\etal{\emph{et al}\onedot}
\makeatother
\usepackage{bm}
\newcommand{\balpha}{\bm{\alpha}}
\newcommand{\bnu}{\bm{\nu}}
\usepackage{mathtools}
\DeclarePairedDelimiterX{\norm}[1]{\lVert}{\rVert}{#1}

\nocopyright

\begin{document}
%
\title{Unsupervised Meta-learning of Figure-Ground Segmentation \\via Imitating Visual Effects}
\author{Ding-Jie  Chen$^{\dag}$, Jui-Ting  Chien$^{\ddag}$, Hwann-Tzong  Chen$^{\ddag}$, and Tyng-Luh Liu$^{\dag}$\\
$^{\dag}$Institute of Information Science, Academia Sinica, Taiwan \\
$^{\ddag}$Department of Computer Science, National Tsing Hua University, Taiwan\\
\{djchen.tw, ydnaandy123\}@gmail.com\,, htchen@cs.nthu.edu.tw\,, liutyng@iis.sinica.edu.tw\\
}
\maketitle
\begin{abstract}
This paper presents a ``learning to learn'' approach to figure-ground image segmentation. By exploring webly-abundant images of specific visual effects, our method can effectively learn the visual-effect internal representations in an unsupervised manner and uses this knowledge to differentiate the figure from the ground in an image. Specifically, we formulate the meta-learning process as a compositional image editing task that learns to imitate a certain visual effect and derive the corresponding internal representation. Such a generative process can help instantiate the underlying figure-ground notion and enables the system to accomplish the intended image segmentation. Whereas existing generative methods are mostly tailored to image synthesis or style transfer, our approach offers a flexible learning mechanism to model a general concept of figure-ground segmentation from unorganized images that have no explicit pixel-level annotations. We validate our approach via extensive experiments on six datasets to demonstrate that the proposed model can be end-to-end trained without ground-truth pixel labeling yet outperforms the existing methods of unsupervised segmentation tasks. 
\end{abstract}

\section{Introduction}
In figure-ground segmentation, the regions of interest are conventionally defined by the provided ground truth, which is usually in the form of pixel-level annotations. Without such supervised information from intensive labeling efforts, it is challenging to teach a system to learn what the figure and the ground should be in each image. To address this issue, we propose an unsupervised meta-learning approach that can simultaneously learn both the {\em figure-ground concept} and the corresponding {\em image segmentation}. 
 
The proposed formulation explores the inherent but often unnoticeable relatedness between performing image segmentation and creating visual effects. In particular, to visually enrich a given image with a special effect often first needs to specify the regions to be emphasized. The procedure corresponds to constructing an internal representation that guides the image editing to operate on the target image regions. For this reason, we name such an internal guidance as the {\em Visual-Effect Representation} (VER) of the image. We observe that for a majority of visual effects, their resulting VER is closely related to image segmentation. Another advantage of focusing on visual-effect images is that such data are abundant from the Internet, while pixel-wise annotating large datasets for image segmentation is time-consuming. However, in practice, we only have access to the visual-effect images, but not the VERs as well as the original images. Taking all these factors into account, we reduce the meta-problem of figure-ground segmentation to predicting the proper VER of a given image for the underlying visual effect. Owing to its data richness from the Internet, the latter task is more suitable for our intention to cast the problem within the unsupervised generative framework.

\begin{figure}[t]
    \centering
    \begin{tabular}{c}
    \includegraphics[width=0.45\textwidth]{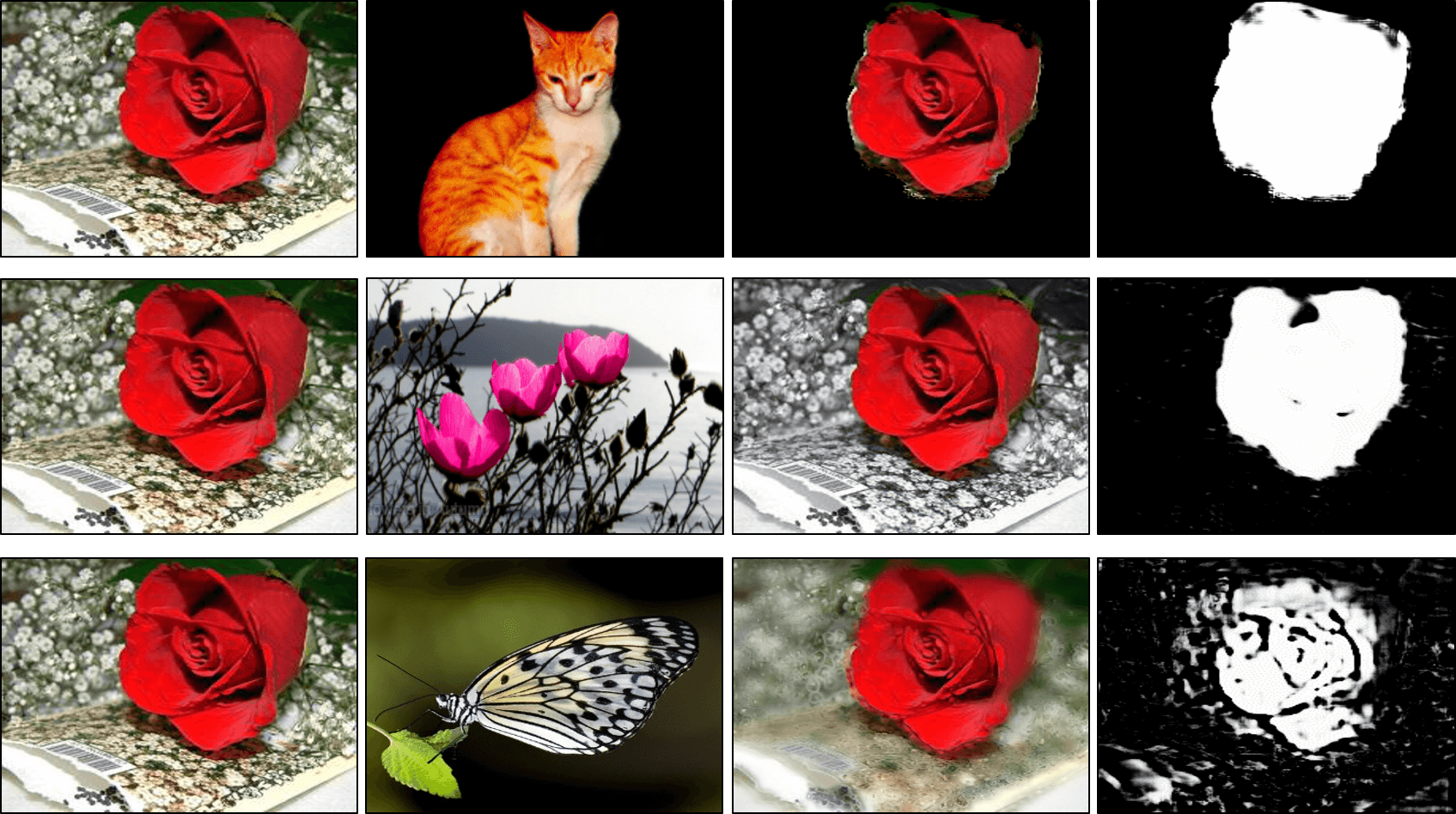}
    \end{tabular}
    \caption{\label{figs:teaser} Given the same image (1st column), imitating different visual effects (2nd column) can yield distinct interpretations of figure-ground segmentation (3rd column), which are derived by our method via referencing the following visual effects (from top to bottom): {\tt black background}, {\tt color selectivo}, and {\tt defocus}/{\tt Bokeh}. The learned VERs are shown in the last column, respectively.}
\end{figure}

\begin{figure*}[!t]
  \centering
  \begin{tabular}{cc}
  \includegraphics[width=0.48\textwidth]{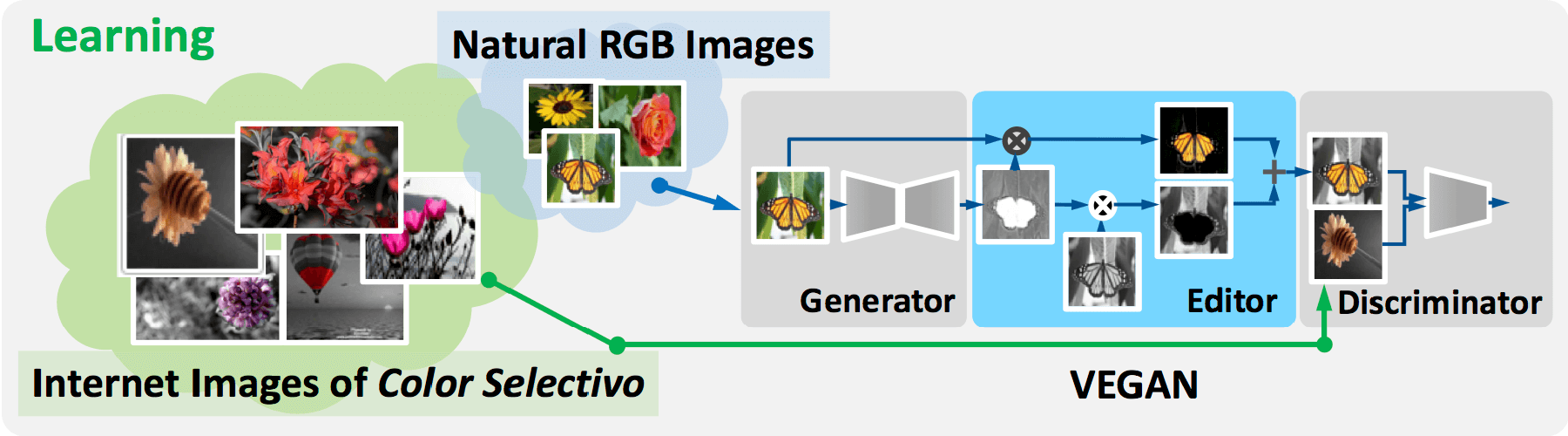} & 
  \includegraphics[width=0.48\textwidth]{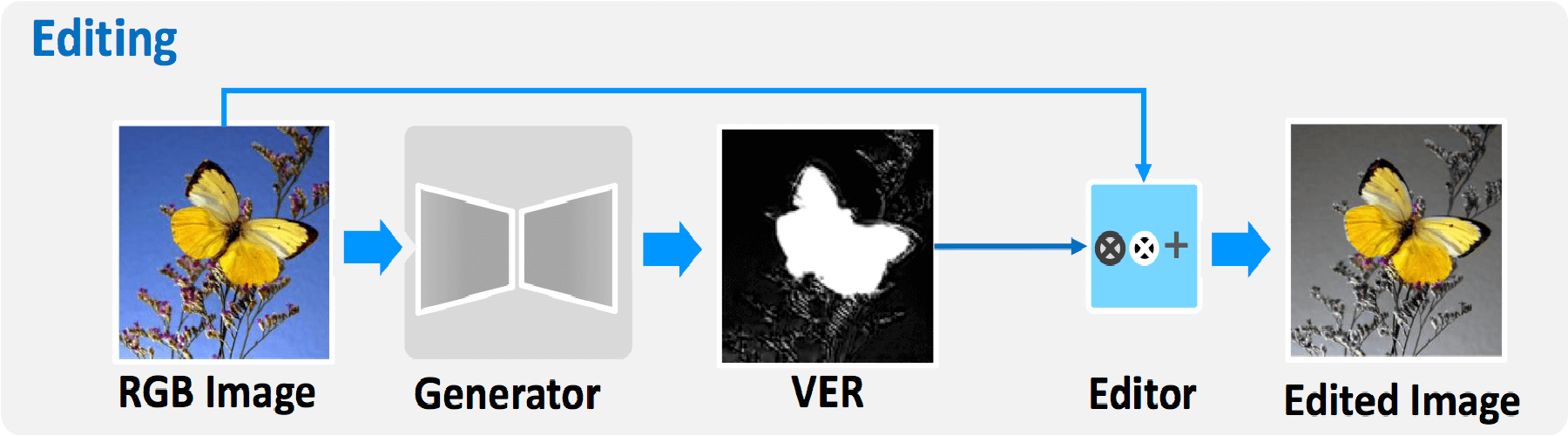} \\
  \end{tabular}
  \caption{Learning and applying our model for the case of ``color selectivo'' visual effect. The image collection for learning is downloaded using Flickr API. Without explicit ground-truth pixel-level annotations being provided, our method can learn to estimate the visual-effect representations (VERs) from \emph{unpaired} sets of natural RGB images and sample images with the expected visual effect. Our generative model is called Visual-Effect GAN (VEGAN), which has an additional component \emph{editor} between the generator and the discriminator. After the unsupervised learning, the generator is able to predict the VER of an input color image for creating the expected visual effect. The VER can be further transformed into figure-ground segmentation. }
  \label{figs:overview}
\end{figure*}

Many compositional image editing tasks have the aforementioned properties. For example, to create the {\tt color selectivo} effect on an image, as shown in Fig.~\ref{figs:overview}, we can \emph{i}) identify the target and partition the image into foreground and background layers, \emph{ii}) convert the color of background layer into grayscale, and \emph{iii}) combine the converted background layer with the original foreground layer to get the final result. The operation of color conversion is local---it simply ``equalizes'' the RGB values of pixels in certain areas. The quality of the result depends on how properly the layers are decomposed. If a part of the target region is partitioned into the background, the result might look less plausible. Unlike the local operations, to localize the proper regions for editing would require certain understanding and analysis of the global or contextual information in the whole image. In this paper, we design a GAN-based model, called Visual-Effect GAN (VEGAN), that can learn to predict the internal representation (\ie, VER) and incorporate such information into facilitating the resulting figure-ground segmentation.

We are thus motivated to formulate the following problem: Given an unaltered RGB image as the input and an image editing task with known compositional process and local operation, we aim to predict the proper VER that guides the editing process to generate the expected visual effect and accomplishes the underlying figure-ground segmentation. We adopt a data-driven setting in which the image editing task is exemplified by a collection of image samples with the expected visual effect. The task, therefore, is to transform the original RGB input image into an output image that exhibits the same effect of the exemplified samples. To make our approach general, we assume that \emph{no} corresponding pairs of input and output images are available in training, and therefore supervised learning is not applicable. That is, the training data does not include pairs of the original color images and the corresponding edited images with visual effects. The flexibility is in line with the fact that although we could fetch a lot of images with certain visual effects over the Internet, we indeed do not know what their original counterpart should look like. Under this problem formulation, several issues are of our interest and need to be addressed. 

First, {\em how do we solve the problem without paired input and output images?} We build on the idea of generative adversarial network and develop a new unsupervised learning mechanism (shown in Figs.~\ref{figs:overview}~\&~\ref{figs:method_network}) to learn the internal representation for creating the visual effect. The \emph{generator} aims to predict the internal VER and the \emph{editor} is to convert the input image into the one that has the expected visual effect. The compositional procedure and local operation are generic and can be implemented as parts of the architecture of a ConvNet. The \emph{discriminator} has to judge the quality of the edited images with respect to a set of sample images that exhibit the same visual effect. The experimental results show that our model works surprisingly well to learn meaningful representation and segmentation without supervision.

Second, {\em where do we acquire the collection of sample images for illustrating the expected visual effect?} Indeed, it would not make sense if we have to manually generate the labor-intensive sample images for demonstrating the expected visual effects. We show that the required sample images can be conveniently collected from the Internet. We provide a couple of scripts to explore the effectiveness of using Internet images for training our model. Notice again that, although the required sample images with visual effects are available on the Internet, their original versions are unknown. Thus supervised learning of pairwise image-to-image translation cannot be applied here.

Third, {\em what can the VER be useful for, in addition to creating visual effects?} We show that, if we are able to choose a suitable visual effect, the learned VER can be used to not only establish the intended figure-ground notion but also derive the image segmentation. More precisely, as in our formulation the visual-effect representation is characterized by a real-valued response map, the result of figure-ground separation can be obtained via binarizing the VER. Therefore, it is legitimate to take the proposed problem of VER prediction as a surrogate for unsupervised image segmentation.

We have tested the following visual effects: \emph{i}) {\tt black background}, which is often caused by using flashlight; \emph{ii}) {\tt color selectivo}, which imposes color highlight on the subject and keeps the background in grayscale; \emph{iii}) {\tt defocus/Bokeh}, which is due to depth of field of camera lens. The second column in Fig.~\ref{figs:teaser} shows the three types of visual effects.
For these tasks our model can be end-to-end trained from scratch in an unsupervised manner using training data that do not have either the ground-truth pixel labeling or the paired images with/without visual effects. While labor-intensive pixel-level segmentations for images are hard to acquire directly via Internet search, images with those three effects are easy to collect from photo-sharing websites, such as Flickr, using related tags.

\section{Related Work}
We discuss below some related work on the topics of generative adversarial networks and image segmentation.

\subsection{Generative Adversarial Networks}
The idea of GAN~\cite{GoodfellowPMXWOCB14} is to generate realistic samples through the adversarial game between generator $G$ and discriminator $D$. GAN becomes popular owing to its ability to achieve unsupervised learning. However, GAN also encounters many problems such as instability and model collapsing. Hence later methods~\cite{RadfordMC15,ArjovskyCB17,GulrajaniAADC17} try to improve GAN in both the aspects of implementation and theory. 
DCGAN~\cite{RadfordMC15} provides a new framework that is more stable and easier to train. WGAN~\cite{ArjovskyCB17} suggests to use Wasserstein distance to measure the loss. WGAN-GP~\cite{GulrajaniAADC17} further improves the way of the Lipschitz constraint being enforced, by replacing weight clipping with gradient penalty.

To reduce the burden of $G$, Denton \etal~\cite{DentonCSF15} use a pyramid structure and Karras \etal~\cite{KarrasALL18} consider a progressive training methodology. Both of them divide the task into smaller sequential steps. In our case, we alleviate the burden of $G$ by incorporating some well-defined image processing operations into the network model, \eg, converting background color into grayscale to simulate the visual effect of {\tt color selectivo}, or blurring the background to create the {\tt Bokeh} effect. 

Computer vision problems may benefit from GAN by including an adversarial loss into, say, a typical CNN model. Many intricate tasks have been shown to gain further improvements after adding adversarial loss, such as shadow detection~\cite{NguyenVZHS17}, saliency detection~\cite{PanCMOTSN17}, and semantic segmentation~\cite{LucCCV16}. However, those training methodologies require paired images (with ground-truth) and hence lack the advantage of unsupervised learning.  
For the applications of modifying photo styles, some methods~\cite{LiuBK17,YiZTG17,ZhuPIE17} can successfully achieve image-to-image style transfer using unpaired data, but their results are limited to subjective evaluation. Moreover, those style-transfer methods cannot be directly applied to the task of unsupervised segmentation. 

Since our model has to identify the category-independent subjects for applying the visual effect without using image-pair relations and ground-truth pixel-level annotations, the problem we aim to address is more general and challenging than those of the aforementioned methods.

\subsection{Image Segmentation}
Most of the existing segmentation methods that are based on deep neural networks (DNNs) to treat the segmentation problem as a pixel-level classification problem~\cite{SimonyanZ14a,LongSD15,HeZRS16}. The impressive performance relies on a large number of high-quality annotations. Unfortunately, collecting high-quality annotations at a large scale is another challenging task since it is exceedingly labor-intensive. As a result, existing datasets just provide limited-class and limited-annotation data for training DNNs. DNN-based segmentation methods thus can only be applied to a limited subset of category-dependent segmentation tasks. 

To reduce the dependency of detailed annotations and to simplify the way of acquiring a sufficient number of training data, a possible solution is to train DNNs in a semi-supervised manner~\cite{HongNH15,SoulySS17} or a weakly-supervised manner~\cite{DaiHS15,KwakHH17,PinheiroC15} with a small number of pixel-level annotations. In contrast, our model is trained without explicit ground-truth annotations.

Existing GAN-based segmentation methods~\cite{NguyenVZHS17,LucCCV16} improve their segmentation performance using mainly the adversarial mechanism of GANs. The ground-truth annotations are needed in their training process for constructing the adversarial loss, and therefore they are GAN-based but not ``unsupervised'' from the perspective of application and problem definition.  

We instead adopt a meta-learning viewpoint to address figure-ground segmentation. Depending on the visual effect to be imitated, the proposed approach interprets the task of image segmentation according to the learned VER. As a result, our model indeed establishes a general setting of figure-ground segmentation, with the additional advantage of generating visual effects or photo-style manipulations.

\begin{figure*}[t]
  \centering
  \includegraphics[width=0.95\textwidth,height=0.21\textheight]{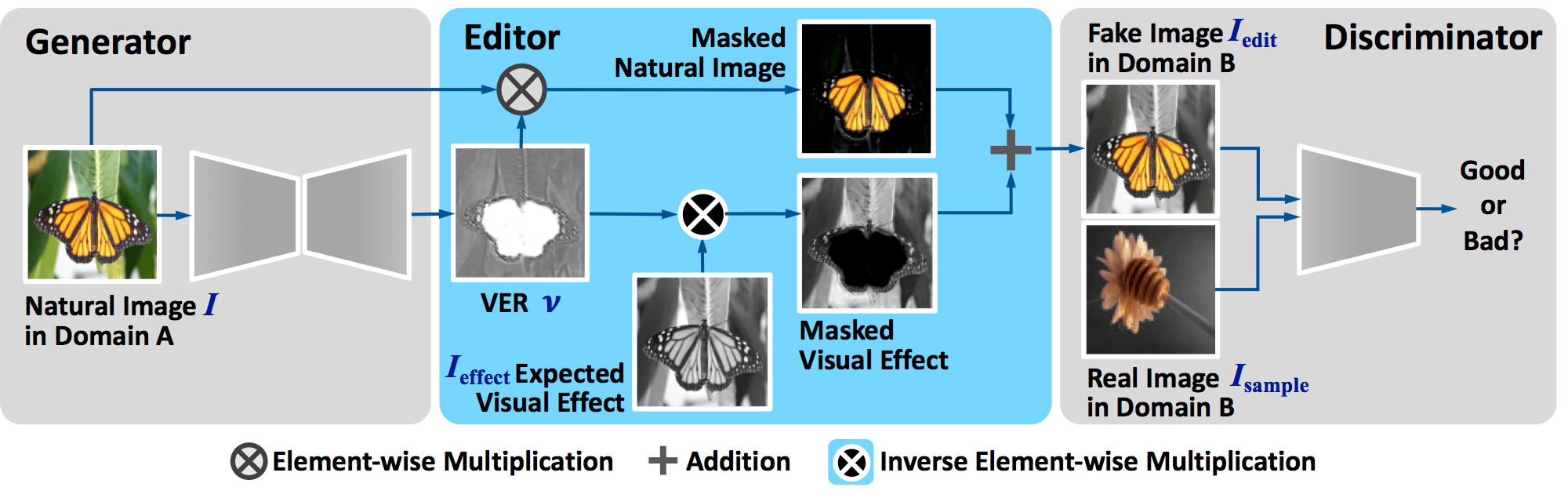}
  \caption{The proposed Visual-Effect GAN (VEGAN) model. Here we take {\tt color selectivo} as the expected visual effect. The visual-effect representation (VER) produced by the generator indicates the strength of the visual effect at each location. The \emph{editor} uses a well-defined trainable procedure (converting RGB to grayscale in this case) to create the expected visual effect. The discriminator receives the edited image $I_\mathrm{edit}$ and evaluates how good it is. To train VEGAN, we need unpaired images from two domains. Domain A comprises real RGB images and Domain B comprises images with the expected visual effect.}
  \label{figs:method_network}
\end{figure*}

\section{Our Method}\label{sec:approach}
Given a natural RGB image $I$ and an expected visual effect with known compositional process and local operation, the proposed VEGAN model learns to predict the visual-effect representation (VER) of $I$ and to generate an edited image $I_\mathrm{edit}$ with the expected effect. Fig.~\ref{figs:overview} illustrates the core idea. The training data are from two \emph{unpaired} sets: the set $\{I\}$ of original RGB images and the set $\{I_\mathrm{sample}\}$ of images with the expected visual effect.
The learning process is carried out as follows: \emph{i}) \emph{Generator} predicts the VER $\bnu$ of the image $I$. \emph{ii}) \emph{Editor} uses the known local operation to create an edited image $I_\mathrm{edit}$ possessing the expected visual effect. \emph{iii}) \emph{Discriminator} judges the quality of the edited images $I_\mathrm{edit}$ with respect to a set $\{I_\mathrm{sample}\}$ of sample images that exhibit the same visual effect. \emph{iv}) \emph{Loss} is computed for updating the whole model. Fig.~\ref{figs:method_network} illustrates the components of VEGAN. Finally, we perform \emph{Binarization} on VER for quantitatively assess the outcome of figure-ground segmentation.

\paragraph{\textbf{Generator}:}
The task of the generator is to predict the VER $\bnu$ that can be used to partition the input image $I$ into foreground and background layers. Our network architecture is adapted from the state-of-the-art methods~\cite{JohnsonAF16,ZhuPIE17} which show impressive results on image style transfer. The architecture follows the rules suggested by DCGAN \cite{RadfordMC15} such as replacing pooling layer with strided convolution. Our base architecture also uses the $9$-residual-blocks version of \cite{JohnsonAF16}. We have also tried a few slightly modified versions of the generator. The differences and details are described in the experiments.

\paragraph{\textbf{Discriminator}:}
The discriminator is trained to judge the quality of the edited images $I_\mathrm{edit}$ with respect to a set $\{I_\mathrm{sample}\}$ of sample images that exhibit the same effect. We adopt a $70 \times 70$ patchGAN \cite{IsolaZZE17,LedigTHCCAATTWS17,LiW16,ZhuPIE17} as our base discriminator network. PatcahGAN brings some benefits with multiple overlapping image patches. Namely, the scores change more smoothly and the training process is more stable. Compared with a full-image discriminator, the receptive field of the $70\times70$ patchGAN might not capture the global context.
In our work, the foreground objects are sensitive to their position in the whole image and are center-biased. If there are several objects in the image, our method would favor to pick out the object closest to the center. In our experiment, $70 \times 70$ patchGAN does produce better segments along the edges, but sometimes the segments tend to be tattered. A full-image discriminator~\cite{GoodfellowPMXWOCB14,RadfordMC15,ArjovskyCB17,GulrajaniAADC17}, on the other hand, could give coarser but more compact and structural segments.

\paragraph{\textbf{Editor}:}
The editor is the core of the proposed model. Given an input image $I$ and its VER $\bnu$ predicted by the generator, the editor is responsible for creating a composed image $I_\mathrm{edit}$ containing the expected visual effect. The first step is based on the well-defined procedure to perform local operations on the image and generate the expected visual effect $I_\mathrm{effect}$. More specifically, in our experiments we define three basic local operations: \emph{black-background}, \emph{color-selectivo}, and \emph{defocus/Bokeh}, which involve \emph{clamping-to-zero}, \emph{grayscale conversion}, and \emph{$11\times11$ average pooling}, respectively. The next step is to combine the edited background layer with the foreground layer to get the final editing result $I_\mathrm{edit}$. An intuitive way is to use the VER $\bnu$ as an alpha map $\balpha$ for image matting, \ie, $I_\mathrm{edit} = \balpha \otimes I + (1-\balpha) \otimes I_\mathrm{effect}$, where $\balpha =\{\alpha_{ij}\}$, $\alpha_{ij} \in (0, 1)$ and $\otimes$ denotes the element-wise multiplication.
However, in our experiments, we find that it is better to have $\bnu =\{\nu_{ij}\}$, $\nu_{ij} \in (-1,1)$ with hyperbolic-tangent as the output. Hence we combine the two layers as follows: 
\begin{equation}\label{eq:method_alpha2}
  	\begin{aligned}	
  	I_\mathrm{edit} =  \tau(\bnu \otimes (I - I_\mathrm{effect}) + I_\mathrm{effect}), \;\; \nu_{ij} \in (-1, 1) \,,
  	\end{aligned} 
\end{equation}  
\noindent where $\tau(\cdot)$ truncates the values to be within $(0, 255)$, which guarantees the $I_\mathrm{edit}$ can be properly rendered. Under this formulation, our model turns to learning the residual.  

\paragraph{\textbf{Loss}:}
We refer to SOTA algorithms~\cite{ArjovskyCB17,GulrajaniAADC17} to design loss functions $\mathcal{L}_G$ and $\mathcal{L}_D$ for generator ($G$) and discriminator ($D$):
\begin{align}
    \mathcal{L}_G &=  - \mathbb{E}_{x \thicksim \mathbb{P}_{g}}[D(x))] \,, \label{eq:gen_lost} \\
    \mathcal{L}_D &= \mathbb{E}_{{x} \thicksim \mathbb{P}_{g}}[D({x}))] - \mathbb{E}_{y \thicksim \mathbb{P}_{r}}[D(y)]\; \nonumber \\
  		&\quad + \;\lambda_{gp}\; \mathbb{E}_{\hat{x} \thicksim \mathbb{P}_{\hat{x}}}[(\norm{\nabla_{\hat{x}}D(\hat{x})}_{2} - 1)^{2}] \,. \label{eq:dis_lost}
\end{align}
\noindent
We alternately update the generator by Eq.~\ref{eq:gen_lost} and the discriminator by Eq.~\ref{eq:dis_lost}. In our formulation, $x$ is the edited image $I_\mathrm{edit}$, $y$ is an image $I_\mathrm{sample}$ which exhibits the expected visual effect, $\mathbb{P}_{g}$ is the edited image distribution, $\mathbb{P}_{r}$ is the sample image distribution, and $\mathbb{P}_{\hat{x}}$ is for sampling uniformly along straight lines between image pairs from $\mathbb{P}_{g}$ and $\mathbb{P}_{r}$. We set the learning rate, $\lambda_{gp}$, and other hyper-parameters the same as the configuration of WGAN-GP \cite{GulrajaniAADC17}. We keep the history of previously generated images and update the discriminator according to the history. We use the same way as~\cite{ZhuPIE17} to store 50 previously generated images $\{I_\mathrm{edit}\}$ in a buffer. The training images are of size $224\times224$, and the batch size is $1$. 

\paragraph{\textbf{Binarization}:}
The VEGAN model can be treated as aiming to predict the strength of the visual effect throughout the whole image. Although the VER provides effective intermediate representation for generating plausible edited images toward some expected visual effects, we observe that sometimes the VER might not be consistent with an object region, particularly with the {\tt Bokeh} effect. Directly thresholding VER to make a binary mask for segmentation evaluation will cause some degree of false positives and degrade the segmentation quality. In general, we expect that the segmentation derived from the visual-effect representation to be smooth within an object and distinct across object boundaries. To respect this observation, we describe, in what follows, an optional procedure to obtain a smoothed VER and enable simple thresholding to yield a good binary mask for quantitative evaluation. Notice that all the VER maps visualized in this paper are obtained without binarization.
 
To begin with, we over-segment \cite{AchantaSSLFS12} an input image $I$ into a superpixel set $\mathcal{S}$ and construct the corresponding superpixel-level graph $\mathcal{G}=(\mathcal{S},\mathcal{E},\mathcal{\omega})$ with the edge set $\mathcal{E}$ and weights $\omega$. Each edge $e_{ij} \in \mathcal{E}$ denotes the spatial adjacency between superpixels $s_i$ and $s_j$. The weighting function $\omega: \mathcal{E} \rightarrow [0,1]$ is defined as $\omega_{ij} = e^{-\theta_1 \| c_i-c_j \|}$, where $c_i$ and $c_j$ respectively denote the CIE Lab mean colors of two adjacent superpixels. Then the weight matrix of the graph is $\mathbf{W}=[\mathcal{\omega}_{ij}]_{|\mathcal{S}| \times |\mathcal{S}|}$.

We then smooth the VER via propagating the averaged value of each superpixel to all other superpixels. To this end, we use $r_i$ to denote the mean VER value of superpixel $s_i$ where $r_i = \frac{1}{|s_i|} \sum_{(i,j) \in s_i} \nu_{ij}$ and $|s_i|$ is the number of pixels within $s_i$. The propagation is carried out according to the feature similarity between every superpixel pair. Given the weight matrix $\mathbf{W}$, the pairwise similarity matrix $\mathbf{A}$ can be constructed as $\mathbf{A} = (\mathbf{D}-\theta_2 \mathbf{W})^{-1}\mathbf{I}$, where $\mathbf{D}$ is a diagonal matrix with each diagonal entry equal to the row sum of $\mathbf{W}$, $\theta_2$ is a parameter in $(0,1]$, and $\mathbf{I}$ is the $|\mathcal{S}|$-by-$|\mathcal{S}|$ identity matrix \cite{ZhouBLWS03}.  Finally, the smoothed VER value of each superpixel can be obtained by 
    \begin{equation}\label{eq:affGlobal}
        [\hat{r}_1, \hat{r}_2, \ldots, \hat{r}_{|\mathcal{S}|}]^T  = \mathbf{D}^{-1}_{\mathbf{A}} \mathbf{A} \cdot [r_1, r_2, \ldots, r_{|\mathcal{S}|}]^T \,,
    \end{equation}
where $\mathbf{D}_{\mathbf{A}}$ is a diagonal matrix with each diagonal entry equal to the corresponding row sum of $\mathbf{A}$, and $\mathbf{D}^{-1}_{\mathbf{A}} \mathbf{A}$ yields the row normalized version of $\mathbf{A}$. From Eq.~\ref{eq:affGlobal}, we see that the smoothed VER value $\hat{r}_i$ is determined by not only neighboring superpixels of $s_i$ but also all other superpixels. 

To obtain the binary mask, we set the average value of $\{\hat{r}_1, \hat{r}_2, \ldots, \hat{r}_{|\mathcal{S}|}\}$ as the threshold for obtaining the corresponding figure-ground segmentation for the input $I$. We set parameters $\theta_1 = 10$ and $\theta_2 = 0.99$ in all the experiments.

\section{Experiments}
We first describe the evaluation metric, the testing datasets, the training data, and the algorithms in comparison. Then, we show the comparison results of the relevant algorithms and our approach. Finally, we present the image segmentation and editing results of our approach. More experimental results can be found in the supplementary material.

\noindent{\textbf{Evaluation Metric}.$\;$}
We adopt the intersection-over-union (IoU) to evaluate the binary mask  derived from the VER. The IoU score, which is defined as $\frac{|P \bigcap Q|}{|P \bigcup Q|}$, where $P$ denotes the machine segmentation and $Q$ denotes the ground-truth segmentation. All algorithms are tested on Intel i7-4770 $3.40$ GHz CPU, 8GB RAM, and NVIDIA Titan X GPU.

\noindent{\textbf{Datasets}.$\;$}
The six datasets are GC50 \cite{RotherKB04}, MSRA500, ECSSD \cite{ShiYXJ16}, Flower17 \cite{NilsbackZ06}, Flower102 \cite{NilsbackZ08}, and CUB200 \cite{WahCUB200_2011}. MSRA500 is a subset of the MSRA10K dataset \cite{ChengMHTH15}, which contains 10{,}000 natural images. We randomly partition MSRA10K into two non-overlapping subsets of 500 and 9{,}500 images to create MSRA500 and MSRA9500 for testing and training, respectively. Their statistics are summarized in Table~\ref{tab:statDataset}. Since these datasets provide pixel-level ground truths, we can compare the consistency between the ground-truth labeling and the derived segmentation of each image for VER-quality assessment. 

\begin{table}[!t]
 \footnotesize
 \begin{center}
 \caption{\label{tab:statDataset} Testing datasets and number of images. }
 \begin{tabular}{|@{\;}c@{\;}|@{\;}c@{\;}|@{\;}c@{\;}|@{\;}c@{\;}|@{\;}c@{\;}|@{\;}c@{\;}|}
     \hline
     GC50 & MSRA500 & ECSSD & Flower17 & Flower102 & CUB200  \\
     \hline \hline
     50 & 500 & 1{,}000 & 1{,}360 & 8{,}189 & 11{,}788   \\
     \hline
 \end{tabular}
 \end{center}
\end{table}

\noindent{\textbf{Training Data}.$\;$}
In training the VEGAN model, we consider using the images from two different sources for comparison. The first image source is MSRA9500 derived from the MSRA10K dataset \cite{ChengMHTH15}. The second image source is Flickr, and we acquire unorganized images for each task as the training data. We examine our model on three kinds of visual effects, namely, {\tt black background}, {\tt color selectivo}, and {\tt defocus}/{\tt Bokeh}. 
\begin{itemize}
\item For MSRA9500 images, we randomly select 4{,}750 images and then apply the three visual effects to yield three groups of \emph{images with visual effects}, \ie,  $\{I_\mathrm{sample}\}$. The other 4{,}750 images are hence the input images $\{I\}$ for the generator to produce the \emph{edited images} $\{I_\mathrm{edit}\}$ later. 
\item For Flickr images, we use ``black background,'' ``color selectivo,'' and ``defocus/Bokeh'' as the three query tags, and then collect 4{,}000 images for each query-tag as the \emph{real images with visual effects}. We randomly download additional 4{,}000 images from Flickr as the images to be edited. 
\end{itemize}

\noindent{\textbf{Algorithms in Comparison}.$\;$}
We quantitatively evaluate the learned VER using the standard segmentation assessment metric (IoU). Our approach is compared with several well-known algorithms, including two semantic segmentation algorithms, three saliency based algorithms, and two bounding-box based algorithms, listed as follows:
ResNet \cite{HeZRS16}, 
VGG16 \cite{SimonyanZ14a},
CA \cite{QinLXW15},
MST \cite{TuHYC16},
GBMR \cite{YangZLRY13},
MilCutS and MilCutG \cite{WuZZLT14},
GrabCut \cite{RotherKB04}.
The two \emph{supervised} semantic segmentation algorithms, ResNet and VGG16, are pre-trained on ILSVRC-2012-CLS \cite{RussakovskyDSKS15} and then fine-tuned on MSRA9500 with ground-truth annotations. The bounding boxes of the two bounding-box based algorithms are initialized around the image borders.

\subsection{Quantitative Evaluation}
\label{sec:exp_quantitative}
The first part of experiment aims to evaluate the segmentation quality of different methods. We first compare several variants of the VEGAN model to choose the best model configuration. Then, we analyze the results of the VEGAN model versus the other state-of-the-art algorithms. 

\begin{figure}[t]
    \centering
    \begin{tabular}{cc}
    \includegraphics[width=0.22\textwidth]{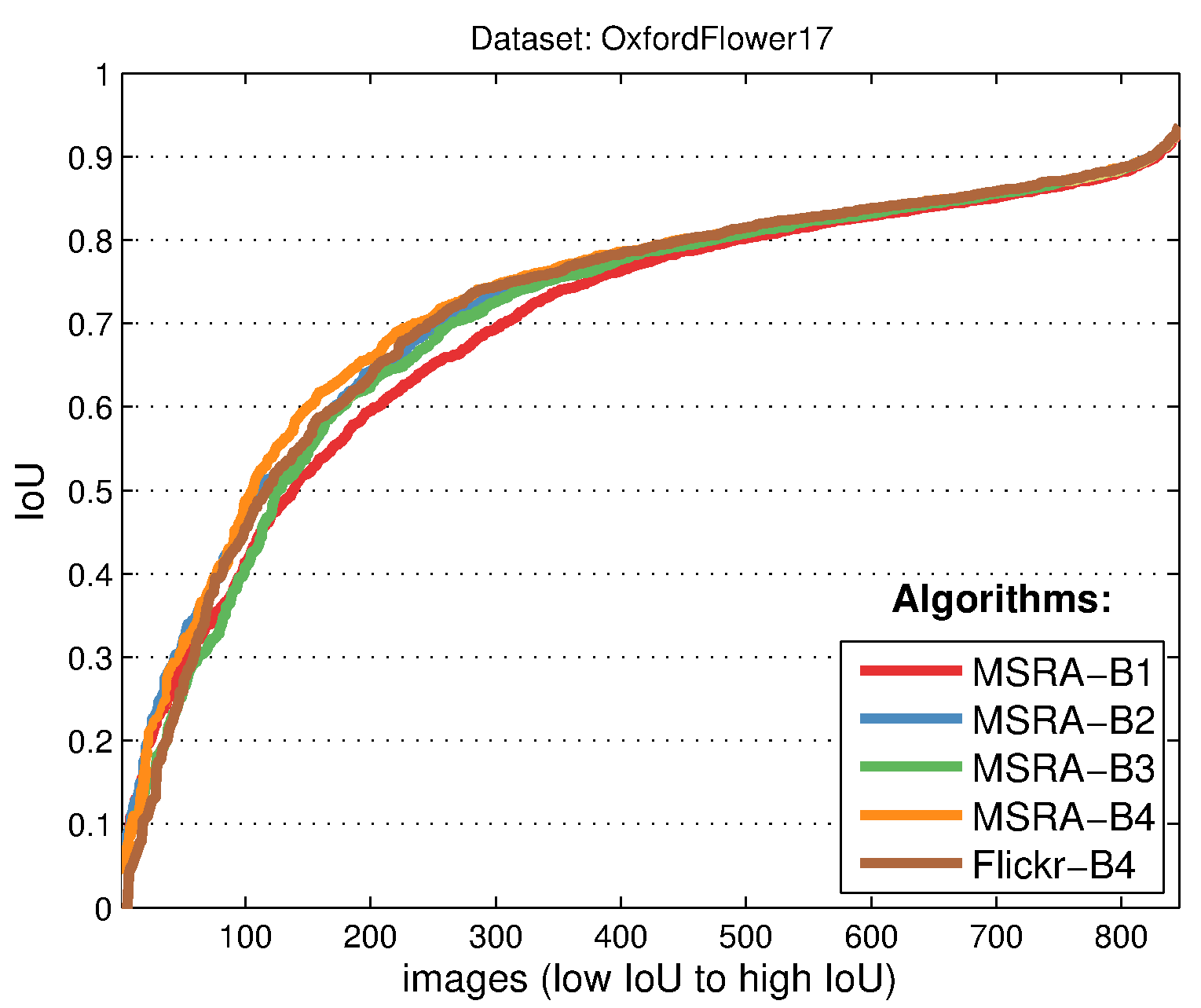}  &
    \includegraphics[width=0.22\textwidth]{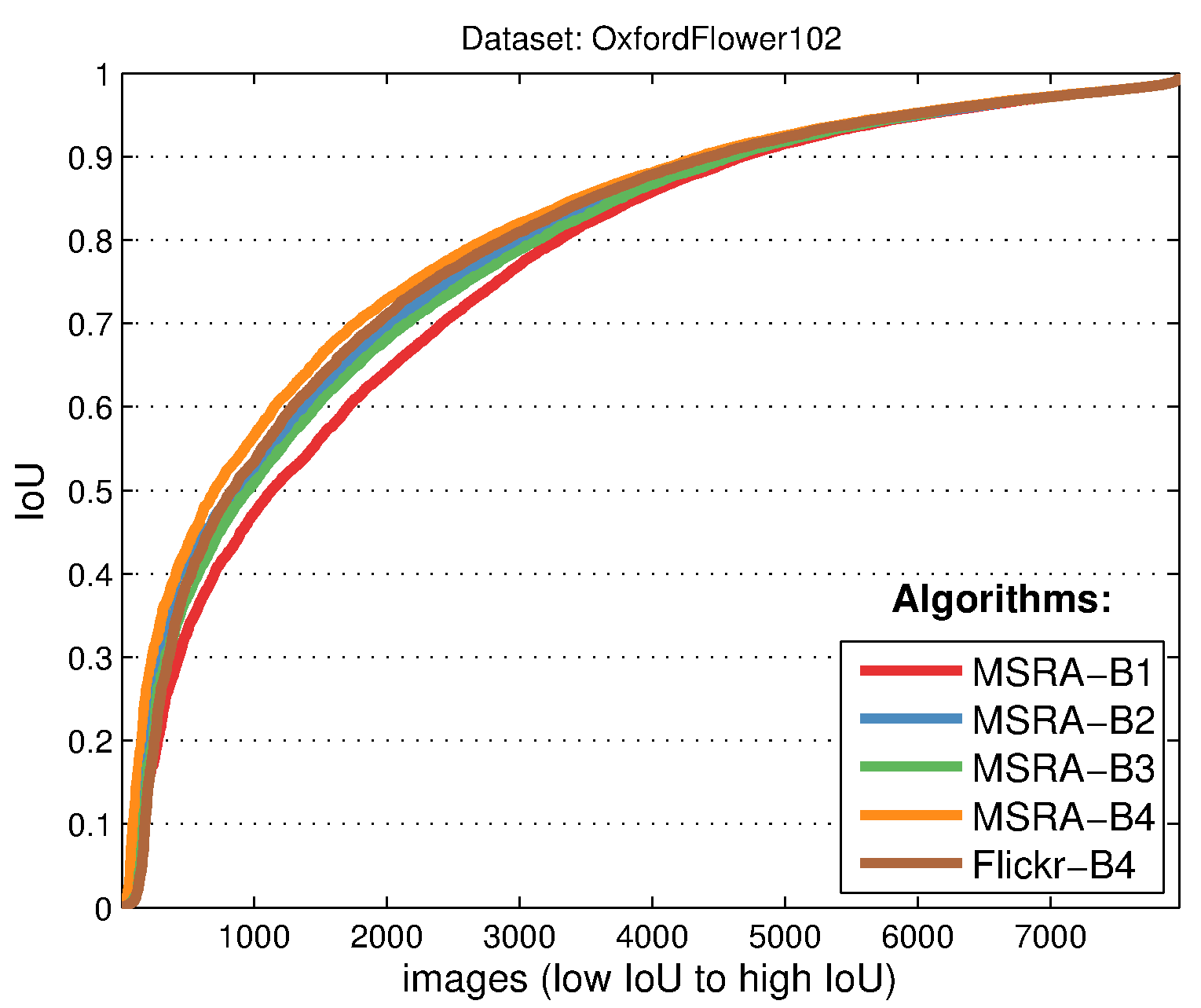} \\
    Testing on Flower17 & Testing on Flower102 
    \end{tabular}
    \caption{\label{figs:comVarFlower} Comparison of two training sources: MSRA9500 and Flickr. Each sub-figure depicts the sorted IoU scores for the variants. Note that Flickr-B4 is trained from the queried Flickr images and performs as well as other variants.}
\end{figure}

\noindent{\textbf{VEGAN Variants}.$\;$}
In the legend blocks of Fig.~\ref{figs:comVarFlower}, we use a compound notation ``\emph{TrainingData} - \emph{Version}'' to account for the variant versions of our model. Specifically, \emph{TrainingData} indicates the image source of the training data. The notation for \emph{Version} contains two characters. 
The first character denotes the type of visual effect:
``B'' for {\tt black background}, ``C'' for {\tt color selectivo}, and  ``D'' for {\tt defocus}/{\tt Bokeh}. 
The second character is the model configuration:
``1'' refers to the combination of base-generator and base-discriminator described in \textbf{Our Method}; 
``2'' refers to using ResNet as the generator; 
``3'' is the model ``1'' with additional skip-layers and replacing transpose convolution with bilinear interpolation; 
``4'' is the model ``3'' yet replacing patch-based discriminator with full-image discriminator.

\begin{table}[t]
  \centering
  \small
  \caption{Comparison of VEGAN variants. All variants are trained with MSRA9500. Each entry shows the \emph{version} and the \emph{mean IoU score} (in parentheses) of a VEGAN variant.}  
  \begin{tabular}{|@{ }c@{ }||@{ }c@{ }|@{ }c@{ }|@{ }c@{ }|@{ }c@{ }|}
  \hline
  \multicolumn{1}{|@{ }c@{ }||}{Visual Effect} & \multicolumn{4}{c|}{Testing Dataset MSRA500 mean IoU}   \\ 
  \hline \hline
    Black Background & B1 (0.67) & B2 (0.73) & B3 (0.70) & \textbf{B4 (0.76)} \\ 
    Color Selectivo  & C1 (0.73) & C2 (0.73) & C3 (0.74) & C4 (0.75) \\ 
    Defocus/Bokeh    & D1 (0.70) & D2 (0.66) & D3 (0.70) & D4 (0.73) \\ \hline 
  \end{tabular}
  \label{tab:varModelMSRA}
\end{table}

We report the results of VEGAN variants in Table~\ref{tab:varModelMSRA}, and depict the sorted IoU scores for the test images in Flower17 and Flower102 datasets in Fig.~\ref{figs:comVarFlower}. It can be seen that all models have similar segmentation qualities no matter what image source is used for training. In Table~\ref{tab:varModelMSRA} and Fig.~\ref{figs:comVarFlower}, the training configuration ``B4'' shows relatively better performance under {\tt black background}. Hence, our VEGAN model adopts the version of MSRA-B4 as a representative variant for comparing with other state-of-the-art algorithms.

\begin{figure}[!t]
    \centering
    \begin{tabular}{cc}
    \includegraphics[width=0.22\textwidth]{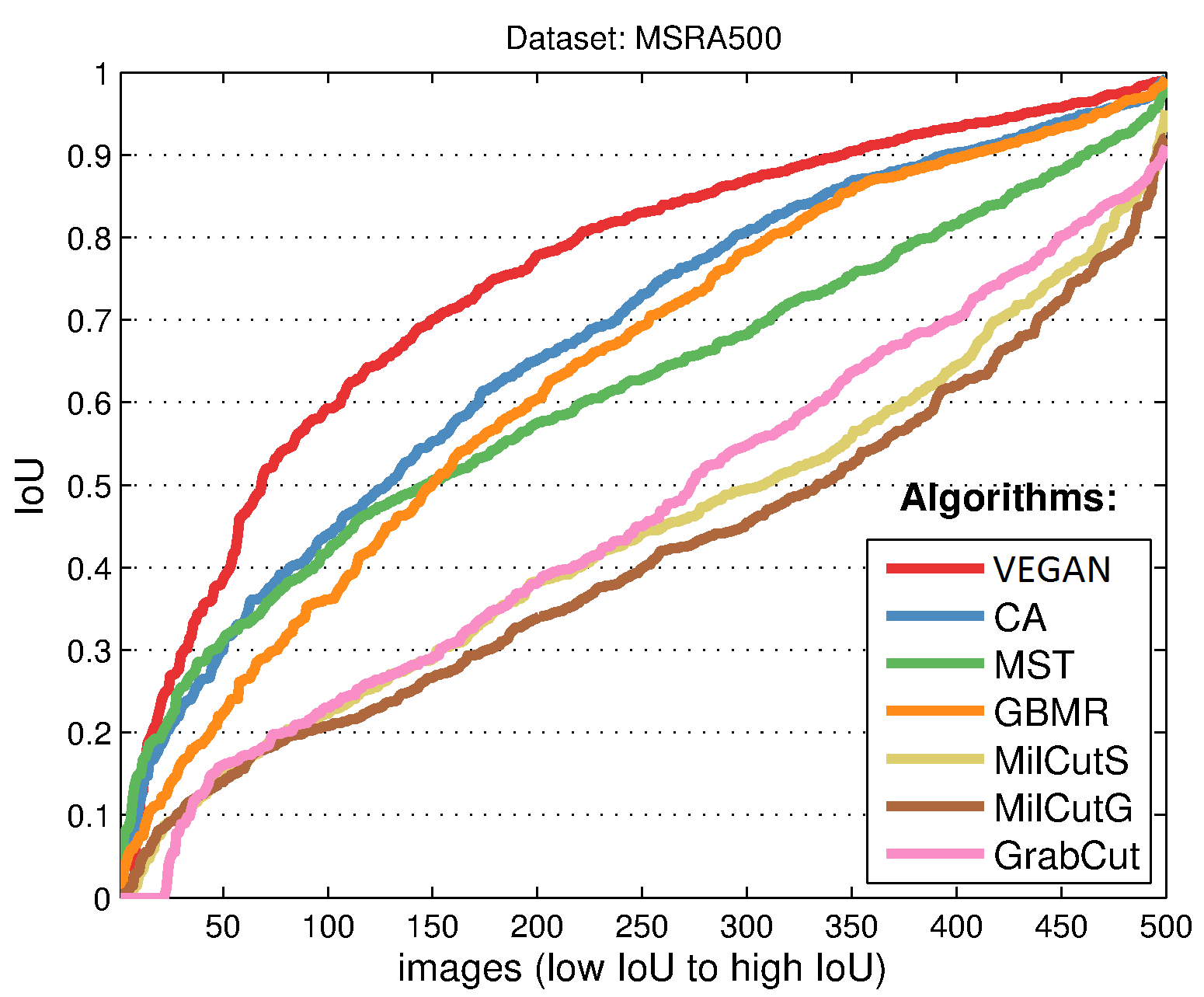} &  
    \includegraphics[width=0.22\textwidth]{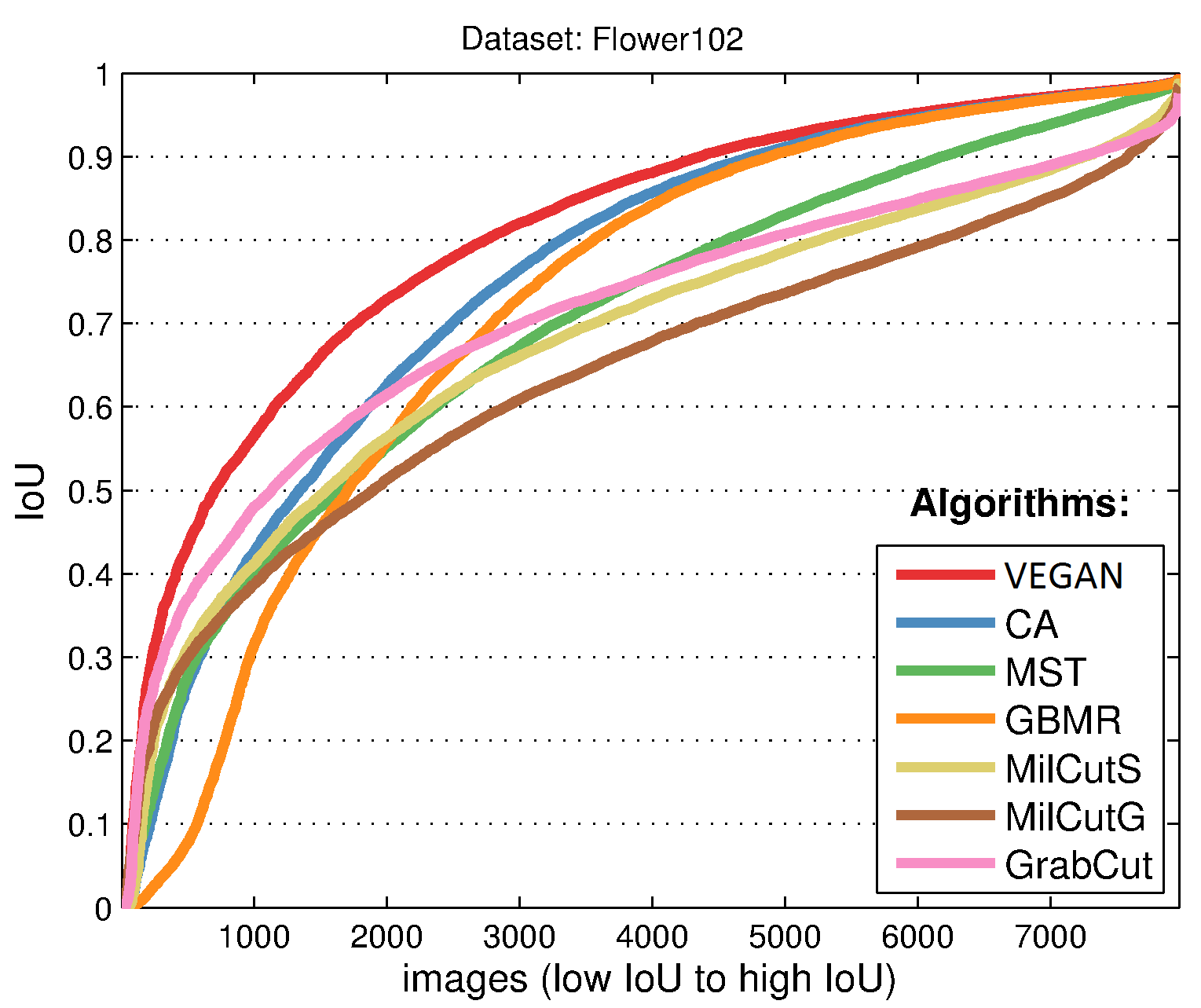}  \\
    MSRA500 & Flower102     \\ 
    \end{tabular}
    \caption{\label{figs:comSOTA_cur} Comparisons among algorithms. Each sub-figure depicts the sorted IoU scores as the segmentation accuracy.}
\end{figure} 

\noindent{\textbf{Unseen Images}.$\;$}
We further analyze the differences of the learned models on dealing with \emph{unseen} and \emph{seen} images. We test the variants B4, C4, and D4 on MSRA500 (unseen) and the subset $\{I\}$ of MSRA9500 (seen). We find that the performance of VEGAN is quite stable. The IoU score for MSRA500 is only $0.01$ lower than the score for MSRA9500 $\{I\}$. Note that, even for the seen images, the ground-truth pixel annotations are unknown to the VEGAN model during training. This result indicates that VEGAN has a good generalization ability to predict segmentation for either seen or unseen images. For comparison, we do the same experiment with the two supervised algorithms, ResNet and VGG16. They are fine-tuned with MSRA9500. The mean IoU scores of ResNet are $0.86$ and $0.94$ for MSRA500 and MSRA9500, respectively. The mean IoU scores of VGG16 are $0.72$ and $0.88$ for MSRA500 and MSRA9500, respectively. The performance of both supervised techniques significantly degrades while dealing with unseen images. 

From the results just described, the final VEGAN model is implemented with the following setting:
\emph{i}) \emph{Generator} uses the $9$-residual-blocks version of \cite{JohnsonAF16}. 
\emph{ii}) \emph{Discriminator} uses the full-image discriminator as WGAN-GP \cite{GulrajaniAADC17}.

\noindent{\textbf{Results}.$\;$}
The top portion of Table~\ref{tab:comSOTA} summarizes the mean IoU score of each algorithm evaluated with the six testing datasets. We first compare our method with five well-known segmentation/saliency-detection techniques, including CA \cite{QinLXW15}, MST \cite{TuHYC16}, GBMR \cite{YangZLRY13}, MilCutS/MilCutG \cite{WuZZLT14}, and GrabCut \cite{RotherKB04}. The proposed VEGAN model outperforms all others on MSRA500, ECSSD, Flower17, and Flower102 datasets, and is only slightly behind the best on GC50 and CUB200 datasets. 
 
The bottom portion of Table~\ref{tab:comSOTA} shows the results of two SOTA supervised learning algorithms on the six testing datasets. Owing to training with the paired images and ground-truths in a ``supervised'' manner, the two models of ResNet and VGG16 undoubtedly achieve good performance so that we treat them as the oracle models. Surprisingly, our unsupervised learning model is comparable with or even slightly better than the supervised learning algorithms on the MSRA500, Flower17, and Flower102 datasets.

Fig.~\ref{figs:comSOTA_cur} depicts the sorted IoU scores, where a larger area under curve means better segmentation quality. VEGAN achieves better segmentation accuracy on the two datasets.

\begin{table}[!t]
\centering

{
\scriptsize
\caption{The mean IoU scores of each algorithm on the six datasets. The VEGAN model uses configuration ``MSRA-B4'' for comparison. Notice that the ``supervised learning algorithms'' ResNet \cite{HeZRS16} and VGG16 \cite{SimonyanZ14a} are pre-trained with ILSVRC-2012-CLS and then fine-tuned with MSRA9500.} 
\label{tab:comSOTA}
\def\arraystretch{1.15}
\begin{tabular}{|@{ }c@{ }|@{ }c@{ }||@{ }c@{ }|@{ }c@{ }|@{ }c@{ }|@{ }c@{ }|@{ }c@{ }|@{ }c@{ }|}
\hline
\multicolumn{2}{|c||@{ }}{\multirow{2}{*}{}}     & \multicolumn{6}{@{ }c@{ }|}{Testing Dataset---Mean IoU Score} \\ \cline{3-8} 
\multicolumn{2}{|c||@{ }}{}                      & GC50 & MSRA500 & ECSSD & Flower17 & Flower102 & CUB200  \\ \hline \hline
\multirow{7}{*}{\rot{~Algorithm}} & VEGAN & 0.58 & \textbf{0.76} & \textbf{0.58} & \textbf{0.72} & \textbf{0.81} & 0.52 \\ \cline{2-8} 
                                  & CA      & 0.59 & 0.67  & 0.50  & \textbf{0.72}  & 0.75  & 0.51  \\ \cline{2-8} 
                                  & MST     & \textbf{0.60} & 0.61  & 0.53  & 0.68  & 0.70  & \textbf{0.54}  \\ \cline{2-8} 
                                  & GBMR    & 0.52 & 0.64  & 0.48  & 0.68  & 0.71  & 0.49  \\ \cline{2-8} 
                                  & MilCutS & 0.54 & 0.43  & 0.48  & 0.67  & 0.67  & 0.43  \\ \cline{2-8} 
                                  & MilCutG & 0.50 & 0.41  & 0.46  & 0.64  & 0.63  & 0.41  \\ \cline{2-8} 
                                  & GrabCut & 0.51 & 0.46  & 0.45  & 0.68  & 0.71  & 0.39  \\ \hline \hline 
\multirow{2}{*}{\rot{~{\small Oracle}}}    & ResNet  & 0.70 & 0.86  & 0.67  & 0.72  & 0.79  & 0.60  \\ \cline{2-8}
                                  & VGG16   & 0.68 & 0.72  & 0.63  & 0.73  & 0.80  & 0.57  \\ \hline                  
\end{tabular}
}

{
\footnotesize
\caption{Percentage of preferring the VEGAN results.}
\label{tab:userStudy}
\begin{tabular}{|c||c|c|}
\hline 
26 participants  & mean    & median   \\ \hline \hline
Black Background & 90.70\% & 95.00\%  \\ \hline
Color Selectivo  & 80.30\% & 80.00\%  \\ \hline
Defocus/Bokeh    & 75.75\% & 80.00\%  \\ \hline
\end{tabular}
}
\end{table}

\subsection{Qualitative Evaluation}\label{sec:exp_qualitative}
Fig.~\ref{figs:exp_results} shows the results generated by our VEGAN model under different configurations. Each triplet of images contains the input image, the visual effect representation (VER), and the edited image. The results in Fig.~\ref{figs:exp_results} demonstrate that VEGAN can generate reasonable figure-ground segmentations and plausible edited images with expected visual effects.

\begin{figure}[!ht]
    \centering
    \begin{tabular}{c@{\quad}c}
    \includegraphics[width=0.21\textwidth,height=0.05\textheight]{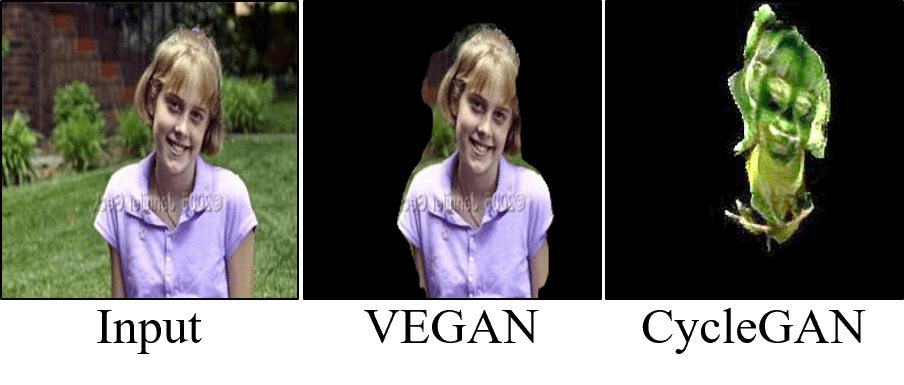} & 
    \includegraphics[width=0.21\textwidth,height=0.05\textheight]{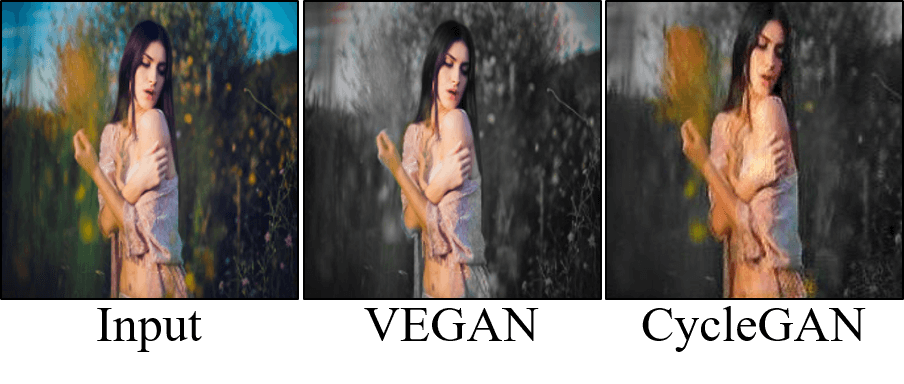} \\
    {\footnotesize Black Background} & {\footnotesize Color Selectivo}   
    \end{tabular}
    \caption{\label{figs:vsCycle} The edited images by VEGAN and CycleGAN. CycleGAN is trained with MSRA9500. The task of imitating {\tt black background} is challenging for CycleGAN.}
\end{figure}

\noindent{\textbf{Visual-Effect Imitation as Style Transfer}.$\;$}
Although existing GAN models cannot be directly applied to learning figure-ground segmentation, some of them are applicable to learning visual-effect transfer, \eg, CycleGAN~\cite{ZhuPIE17}. We use the two sets $\{I\}$ and $\{I_\mathrm{sample}\}$ of MSRA9500 to train CycleGAN, and show some comparison results in Fig.~\ref{figs:vsCycle}. We find that the task of imitating {\tt black background} turns out to be challenging for CycleGAN since the information in $\{I_\mathrm{sample}\}$ is too limited to derive the inverse mapping back to $\{I\}$. Moreover, CycleGAN focuses more on learning the mapping between local properties such as color or texture rather than learning how to create a globally consistent visual effect. VEGAN instead  follows a systematic learning procedure to imitate the visual effect. The generator must produce a meaningful VER so that the editor can compose a plausible visual-effect image that does not contain noticeable artifacts for the discriminator to identify.

\begin{figure}[!ht]
    \centering
    \begin{tabular}{c}
    \includegraphics[width=0.45\textwidth,height=0.32\textheight]{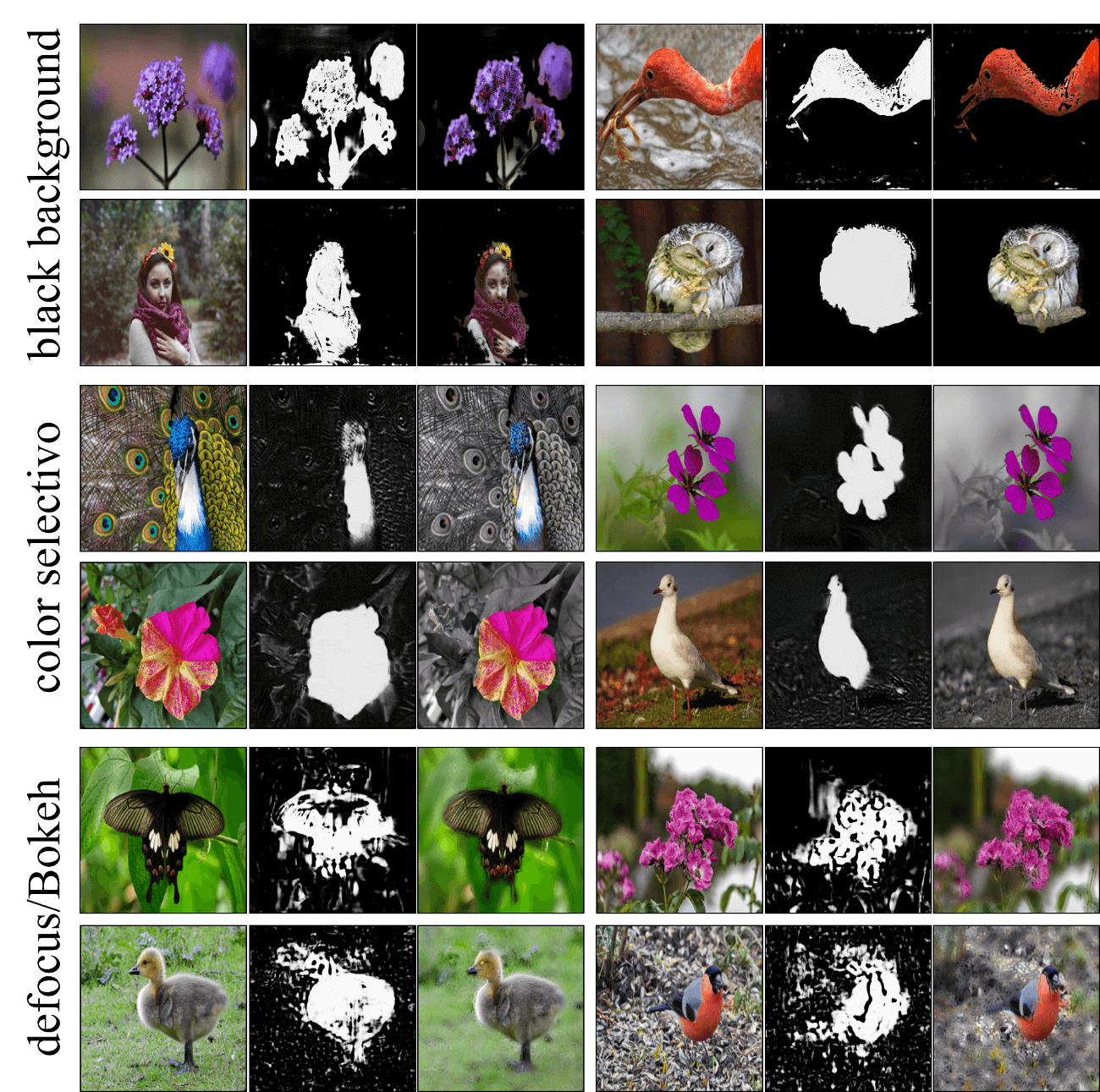} \\
    \end{tabular}
    \caption{\label{figs:exp_results} The edited images generated by VEGAN with respect to specific visual effects. Each image triplet from left to right: the input image, the VER, and the edited image.}
\end{figure}   
    
\begin{figure}[!h]
    \centering
    \begin{tabular}{c}
    \includegraphics[width=0.43\textwidth,height=0.1\textheight]{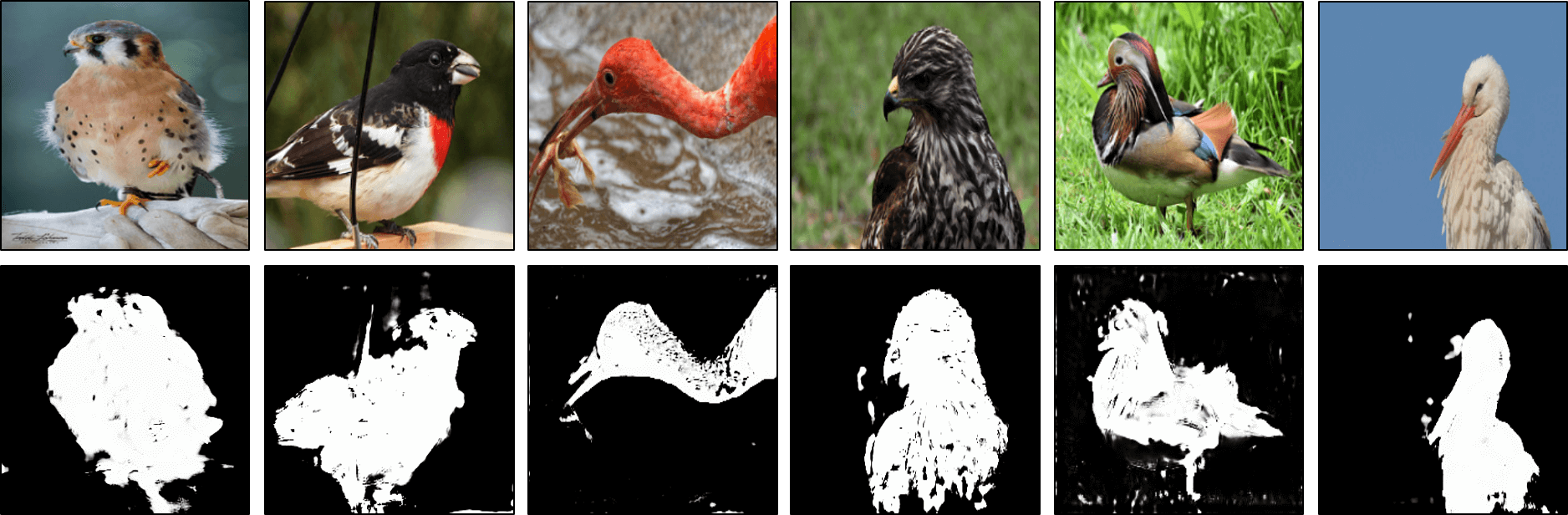}
    \end{tabular}
    \caption{\label{figs:unseen} Testing on Flickr ``bird'' images using VEGAN model trained with Flickr ``flower'' images. The first row shows the input images. The second rows shows the VERs.}
\end{figure}

\noindent{\textbf{User Study}.$\;$}
Table~\ref{tab:userStudy} lists the results from 26 participants. The survey is presented in Google forms, comprising 40 edited image pairs (\eg, as in Fig.~\ref{figs:vsCycle}) for each visual effect. Each user is asked to select the preferred one from an image pair of random order by CycleGAN and VEGAN. 

\noindent{\textbf{Unseen Figures}.$\;$}
To demonstrate that the VEGAN model can learn the general concept of figure-ground segmentation and thus handle unseen foreground ``figures.'' Fig.~\ref{figs:unseen} shows VERs that testing on Flickr ``bird'' images using VEGAN models trained merely with Flick ``flower'' images. The results suggest that the meta-learning mechanism enables VEGAN to identify unseen foreground figures based on the learned knowledge embodied in the generated VERs.

\section{Conclusion} 
We characterize the two main contributions of our method as follows. First, we establish a meta-learning framework to learn a general concept of figure-ground application and an effective approach to the segmentation task. Second, we propose to cast the meta-learning 
as imitating relevant visual effects and develop a novel VEGAN model with following advantages:
\emph{i}) Our model offers a new way to predict meaningful figure-ground segmentation from unorganized images that have no explicit pixel-level annotations. 
\emph{ii}) The training images are easy to collect from photo-sharing websites using related tags. 
\emph{iii}) The \emph{editor} between the generator and the discriminator enables VEGAN to decouple the compositional process of imitating visual effects and hence allows VEGAN to effectively learn the underlying representation (VER) for deriving figure-ground segmentation.  
We have tested three visual effects, including ``black background,'' ``color selectivo,'' and ``defocus/Bokeh'' with extensive experiments on six datasets. For these visual effects, VEGAN can be end-to-end trained from scratch using unpaired training images that have no ground-truth labeling. 

{\scriptsize \noindent {\bf Acknowledgement}. This work was supported in part by the MOST, Taiwan under Grants 107-2634-F-001-002 and 106-2221-E-007-080-MY3.}

\bibliographystyle{aaai}
\bibliography{aaai19.bib}

\subsection*{Appendix A: Quantitative Comparison with SOTA}
We compare the proposed VEGAN model with five well-known segmentation/saliency-detection algorithms, including CA \cite{QinLXW15}, MST \cite{TuHYC16}, GBMR \cite{YangZLRY13}, MilCutS/MilCutG \cite{WuZZLT14}, and GrabCut \cite{RotherKB04}.  

Each sub-figure in Fig.~\ref{figs:comSOTA_cur} depicts the sorted IoU scores. In Fig.~\ref{figs:comSOTA_cur}, a larger area under curve means better segmentation quality. Our VEGAN model has better segmentation accuracy on MSRA500, ECSSD, Flower17, and Flower102 datasets, and is on par with CA \cite{QinLXW15} on Flower17 dataset. Our model is comparable with CA and MST \cite{TuHYC16} on CUB200 dataset, especially for the high-quality segmentation results (when IoU$>$0.7).

\begin{figure*}[!ht]
    \centering
    \begin{tabular}{cc}
    \includegraphics[width=0.43\textwidth]{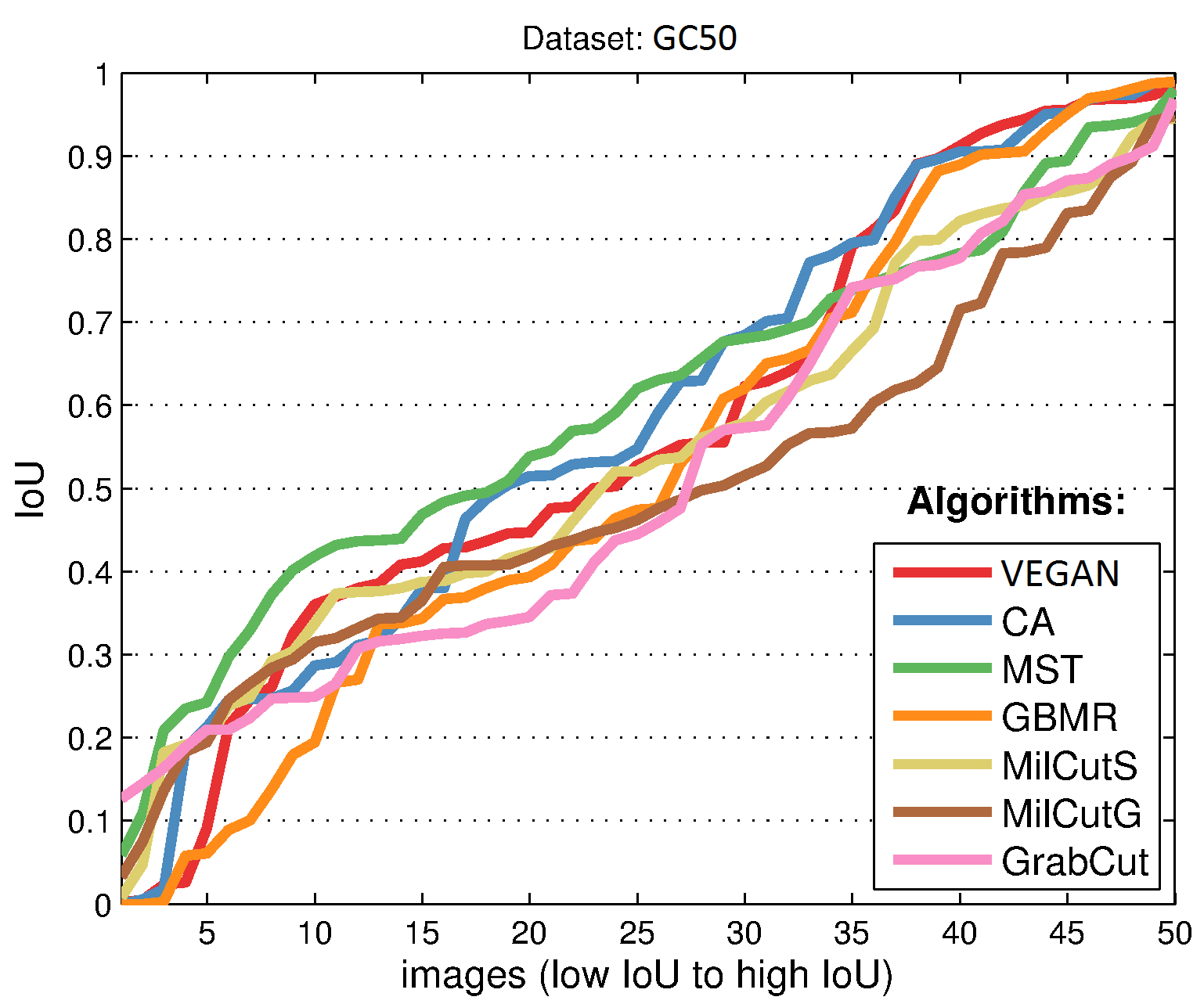} &
    \includegraphics[width=0.43\textwidth]{f_comSOTA_MSRA500-IoU.png} \\  
    GC50 & MSRA500 \\
    \includegraphics[width=0.43\textwidth]{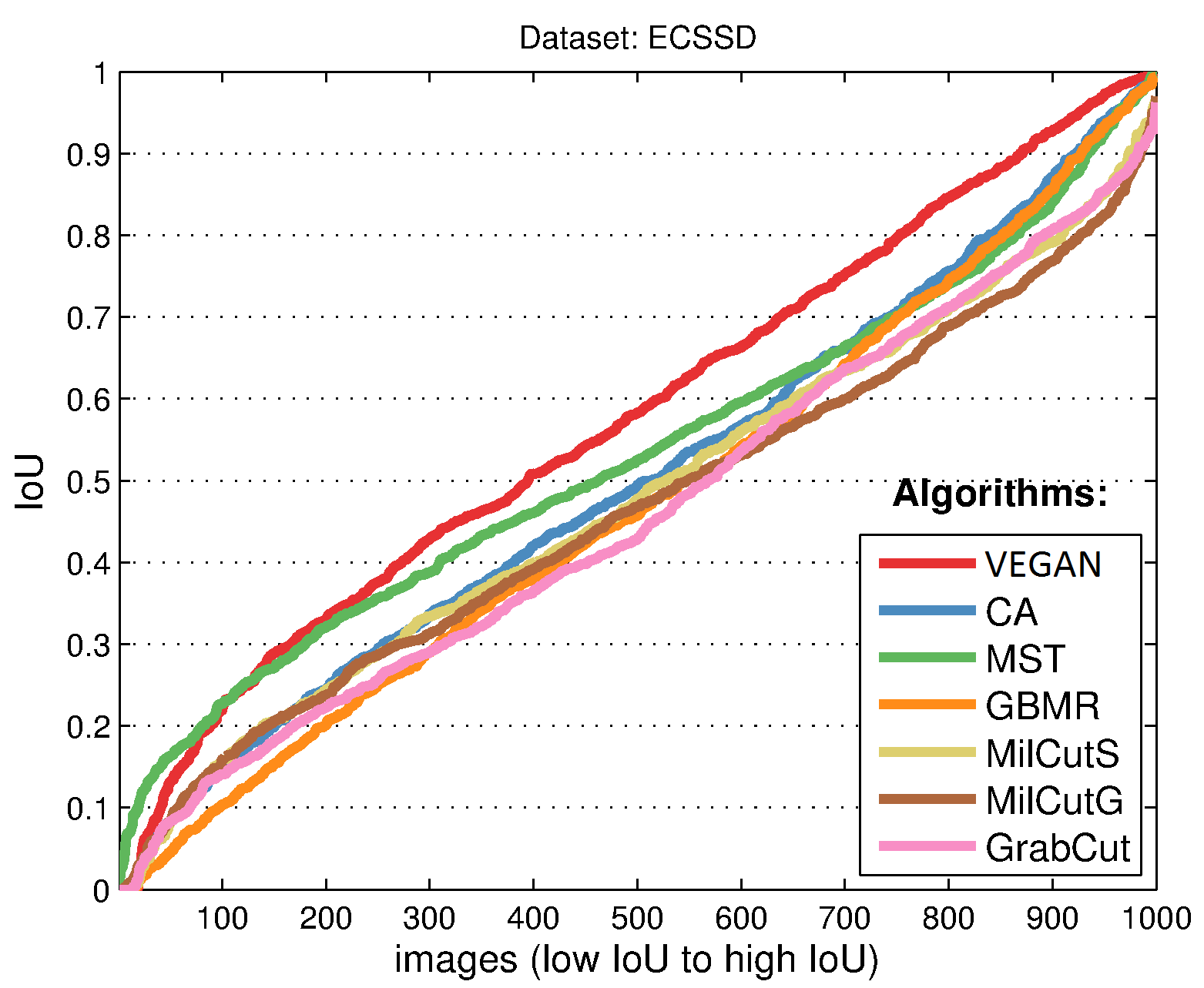}  & 
    \includegraphics[width=0.43\textwidth]{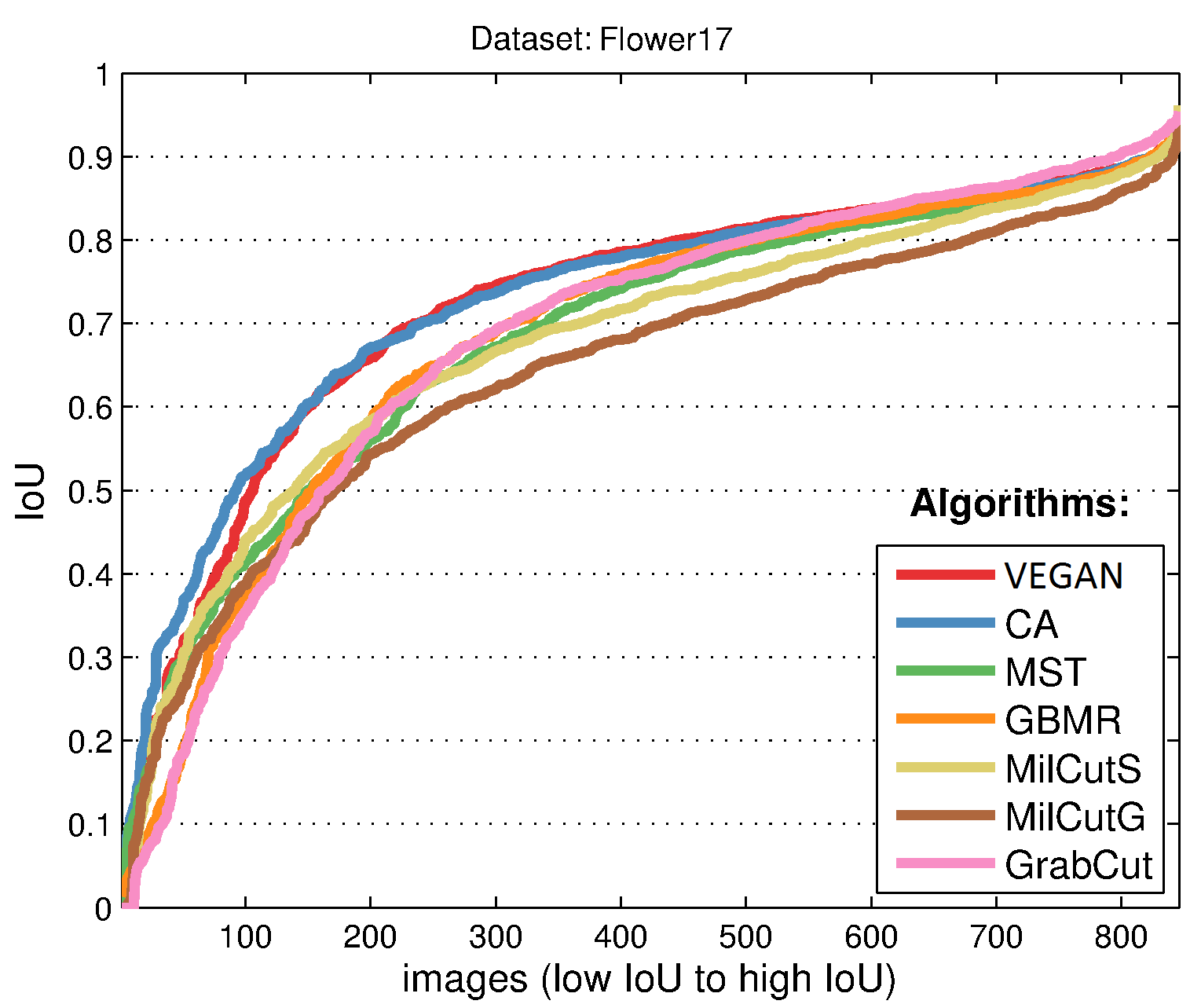} \\
     ECSSD & Flower17\\         
    \includegraphics[width=0.43\textwidth]{f_comSOTA_OxfordFlower102-IoU.png} &  
    \includegraphics[width=0.43\textwidth]{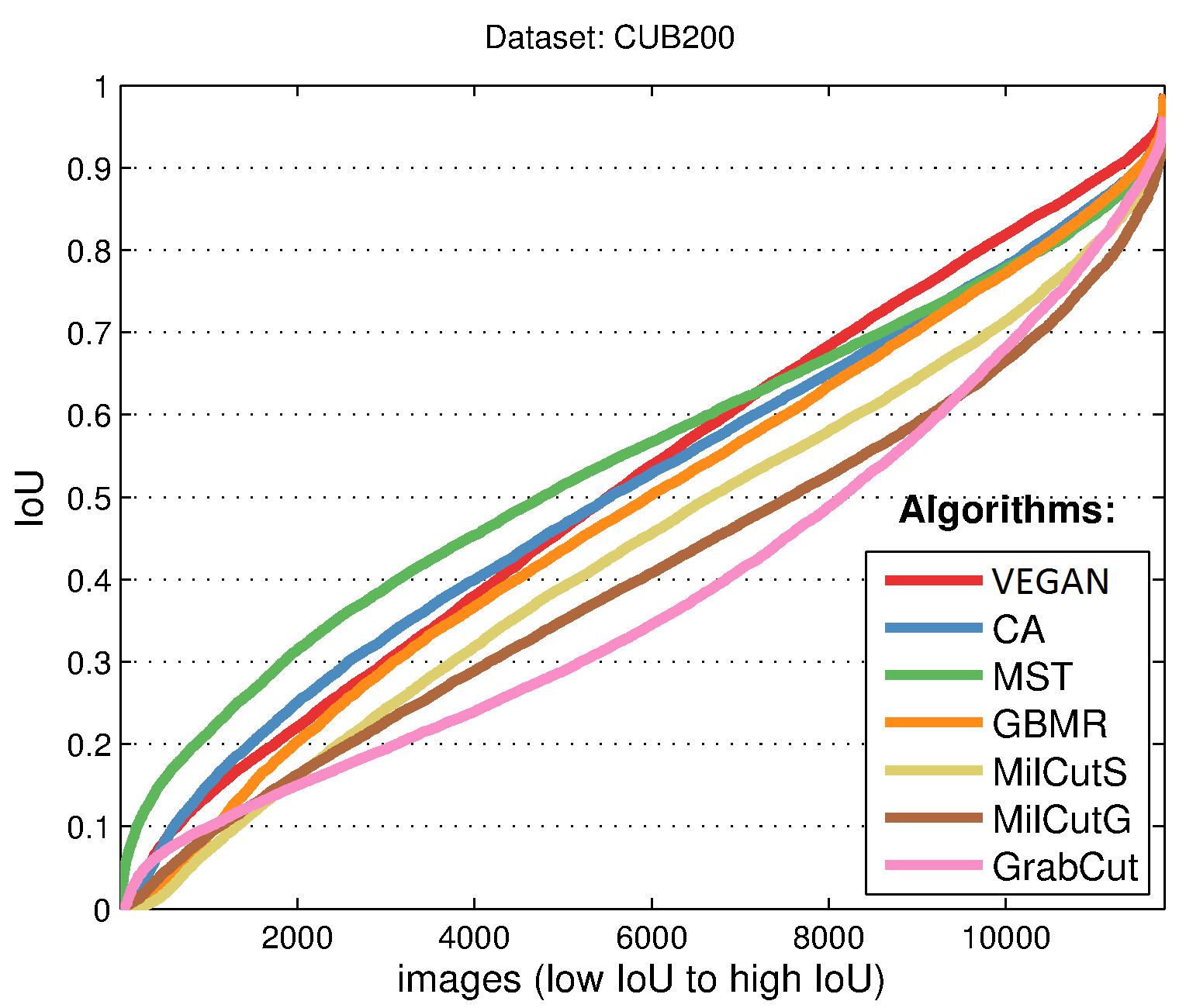}     \\
    Flower102 & CUB200   
    \end{tabular}
    \caption{\label{figs:comSOTA_cur} Comparisons with well-known algorithms, including CA \cite{QinLXW15}, MST \cite{TuHYC16}, GBMR \cite{YangZLRY13}, MilCutS/MilCutG \cite{WuZZLT14}, and GrabCut \cite{RotherKB04}. Each sub-figure depicts the sorted IoU scores as the segmentation accuracy.}
\end{figure*}

\vspace{0.5cm}
\subsection*{Appendix B: Qualitative Results}
Fig.~\ref{figs:exp_results} shows the qualitative results generated by VEGAN's meta-learning process under different configurations. Each triplet of images contains the input image, the VER, and the edited image. The results demonstrate that the VEGAN models can derive distinct interpretations of figure-ground segmentation and generate plausible edited images with expected visual effects.

\
 
\begin{figure*}[!p]
    \centering
    \begin{tabular}{c}
    \includegraphics[width=0.85\textwidth]{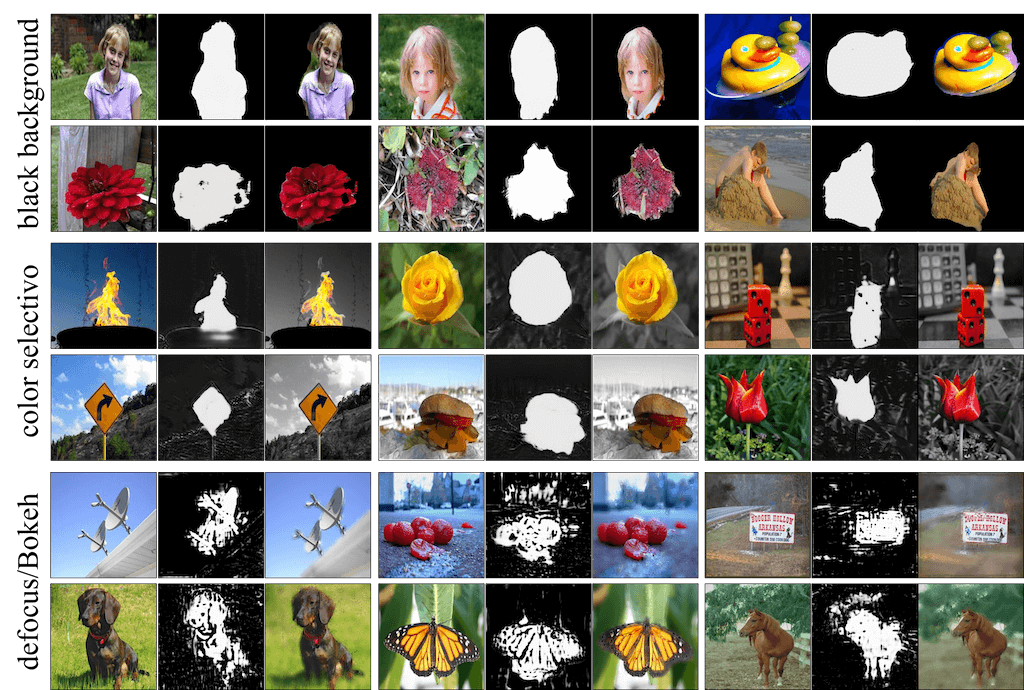} \\
    Testing on MSRA500 using VEGAN models MSRA-B4, MSRA-C4, and MSRA-D4. \\
    \includegraphics[width=0.85\textwidth]{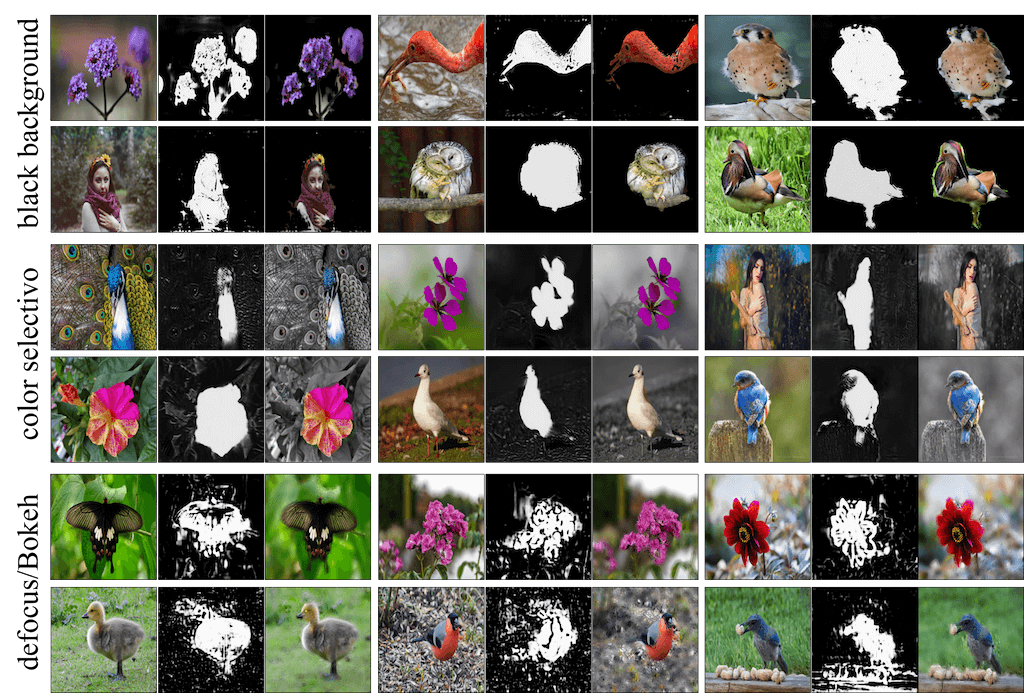} \\
    Testing on Flickr images using VEGAN models Flickr-B4, Flickr-C4, and Flickr-D4. \vspace{-3mm}  
    \end{tabular}
    \caption{\label{figs:exp_results} The edited images generated by our VEGAN models with respect to some expected visual effects. Each image triplet from left to right: the input image, the VER, and the edited image. }
\end{figure*}

\vspace{0.5cm}
\subsection*{Appendix C: VEGAN Variants}
Table~\ref{tab:modelVariants} summarizes the detailed configurations of VEGAN variants. A comparison of these variants is shown in Table~\ref{tab:varModelMSRA}. The performances of B4, C4, and D4 for the three visual effects are all good and do not differ much. Note that, although training with the `black background' visual effect under the setting B4 achieves the best mean IoU score among all variants, training with the visual effect of `color selectivo' performs comparably and consistently well under all of the four settings C1-C4. 

\begin{table*}[ht]
\centering
\caption{\label{tab:modelVariants} The variants of VEGAN. For each visual effect, VEGAN has four versions of configurations. The selected visual effects are {\tt black background} (B), {\tt color selectivo} (C), and {\tt defocus/Bokeh} (D). `\dag' refers to \cite{JohnsonAF16}; `\ddag' refers to \cite{ZhuPIE17}; `$\natural$' refers to \cite{GulrajaniAADC17}. }   
  \begin{tabular}{|c|c|c|c|c|c|}
  \hline
  \multirow{5}{*}{ \bf{\makecell{Per\\Visual Effect \\B, C, or D}} }  
      & \bf{Version}        & \bf{Generator} & \bf{Discriminator} 
      & \bf{Skip-layers} & \bf{Upsampling} \\
  \cline{2-6} \cline{2-6}
      {}  & 1    & the $9$-residual-blocks version \dag & patchGAN \ddag 
          & no  & transposed conv. \\
  \cline{2-6}
      {}  & 2    & ResNet pre-trained & patchGAN \ddag 
          & no  & transposed conv. \\
  \cline{2-6}
      {}  & 3    & the $9$-residual-blocks version \dag & patchGAN \ddag   
          & yes  & bilinear \\
  \cline{2-6}
      {}  & 4    & the $9$-residual-blocks version \dag & WGAN-GP $\natural$
          & yes  & bilinear \\
  \hline
  \end{tabular}
\end{table*}

\begin{table*}[ht]
  \centering
  \caption{\label{tab:varModelMSRA}Comparison of VEGAN variants.  All the variants are trained with MSRA9500 dataset and tested on MSRA500 dataset. Each entry shows the \emph{version} and the \emph{mean IoU score} (in parentheses) of a VEGAN variant. }  
  \begin{tabular}{|c|c||c|c|c|c|}
  \hline
  \multicolumn{2}{|c||}{VEGAN} & \multicolumn{4}{c|}{Testing Dataset MSRA500 mean IoU}            \\ 
  \hline \hline
  \multirow{3}{*}{\makecell{Visual\\Effect}}& Black Background & B1 (0.67) & B2 (0.73) & B3 (0.70) & B4 (0.76) \\ 
                                & Color Selectivo  & C1 (0.73) & C2 (0.73) & C3 (0.74) & C4 (0.75) \\ 
                                & Defocus/Bokeh    & D1 (0.70) & D2 (0.66) & D3 (0.70) & D4 (0.73) \\ \hline 
  \end{tabular}
\end{table*}

\begin{figure}[!h]
    \centering
    \includegraphics[width=0.38\textwidth]{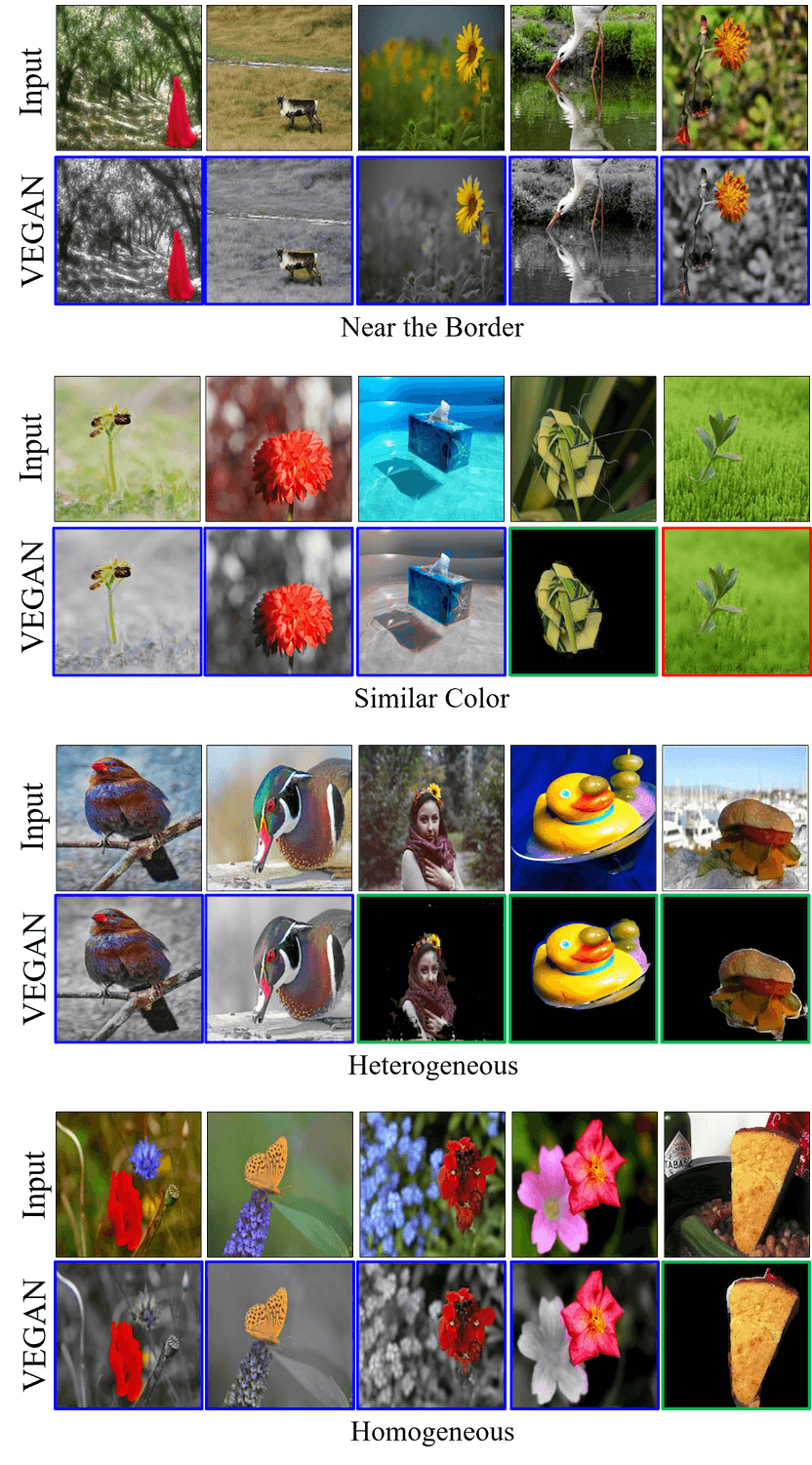} \\ 
    \vspace{-4mm}
    \caption{\label{figs:property} VEGAN can localize the objects that are near the image border or have a similar color distribution as the background. It performs well for either heterogeneous or homogeneous color distributions. Visual effects: {\tt Color selectivo} (blue); {\tt Black background} (green); {\tt Defocus/Bokeh} (red).}
    \vspace{-2mm}
\end{figure}

\vspace{0.5cm}
\subsection*{Appendix D: Versatility of VEGAN}

Our collected Flickr images with the editing effects suggest that people tend to centralize the foreground object. Nevertheless, as shown in Fig.~\ref{figs:property}, the resulting VEGAN model is actually quite versatile. It can localize those object regions that locate near the image border; it can extract foreground regions that have a similar color distribution as the background; it also performs well for either heterogeneous or homogeneous color distributions.

\vspace{0.5cm}
\subsection*{Appendix E: Qualitative Comparison with CycleGAN}

Because state-of-the-art GAN models, e.g., CycleGAN~\cite{ZhuPIE17}, are not explicitly designed for unsupervised learning of figure-ground segmentation, we simply conduct qualitative comparisons with CycleGAN~\cite{ZhuPIE17} on the task of visual-effect transfer rather than the task of figure-ground segmentation. The task of visual-effect transfer is to convert an RGB image into an edited image with the intended visual effect. 

\begin{figure*}[t]
    \centering
    \begin{tabular}{c}
    \vspace{-2mm}
    \includegraphics[width=0.8\textwidth]{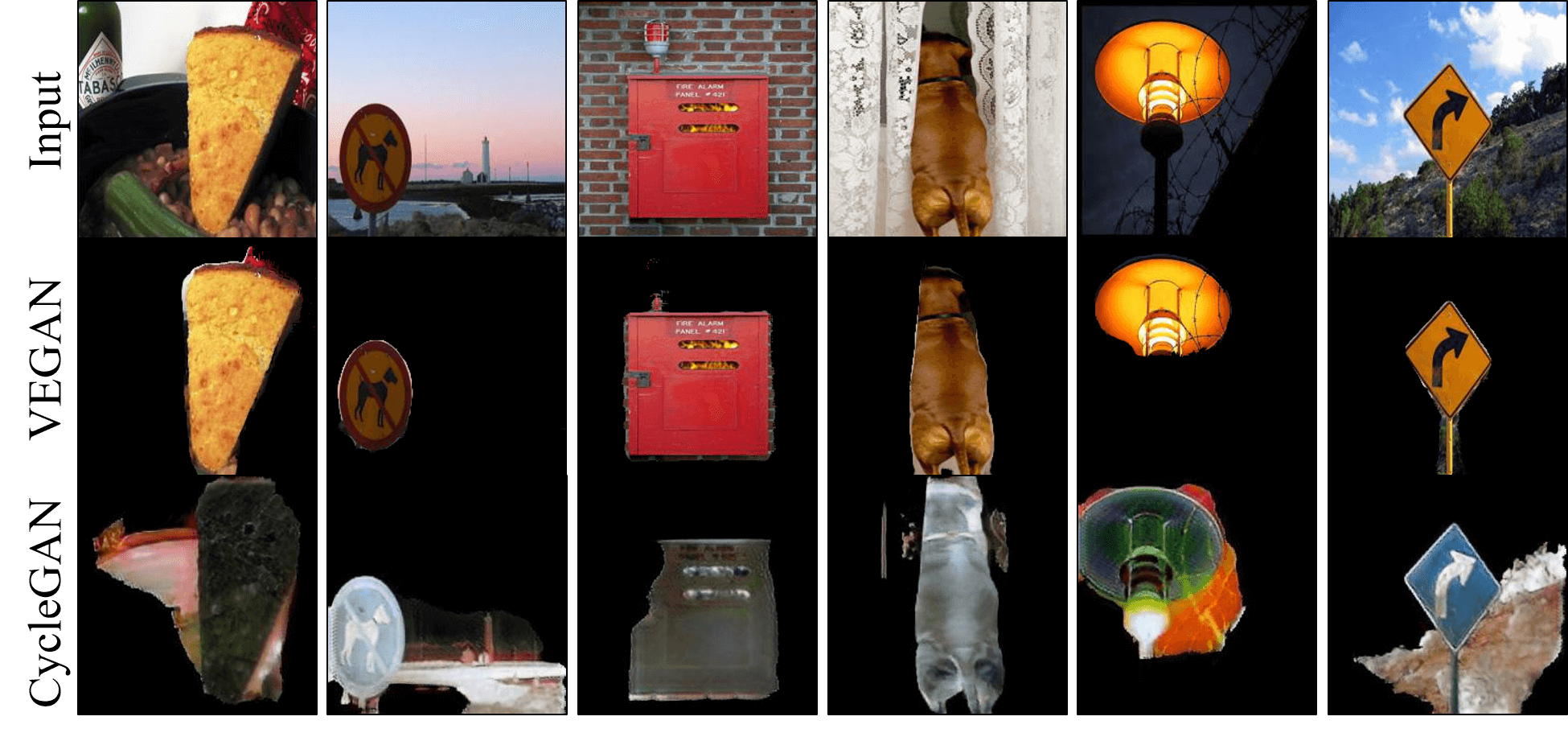} \\ 
    \vspace{3mm}
    `Black background' visual effect  generated by VEGAN (MSRA-B4) and CycleGAN. \\
    \vspace{-2mm}
    \includegraphics[width=0.8\textwidth]{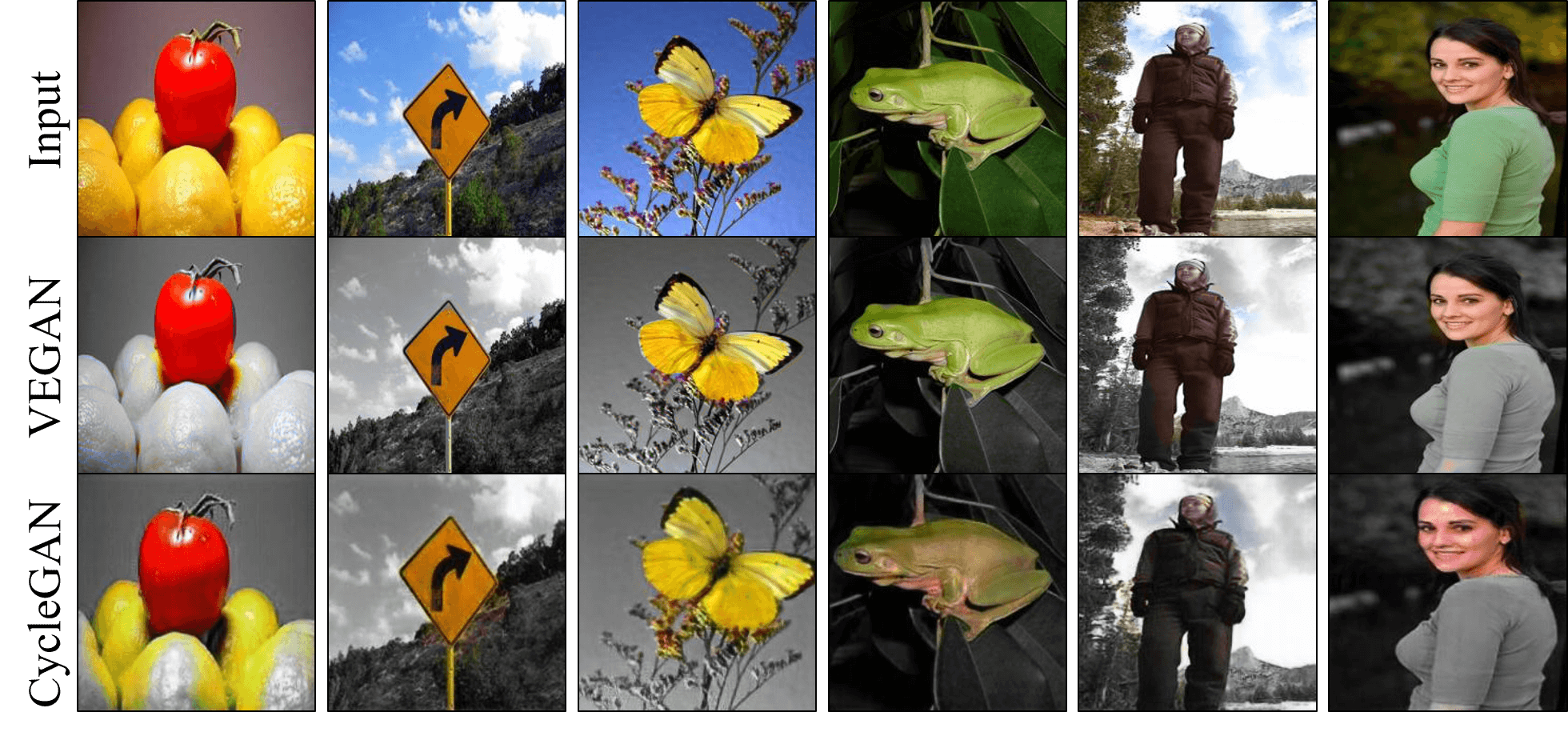} \\
    \vspace{3mm}
    `Color selectivo' visual effect generated by VEGAN (MSRA-C4) and CycleGAN.  \\   
    \vspace{-2mm}
    \includegraphics[width=0.8\textwidth]{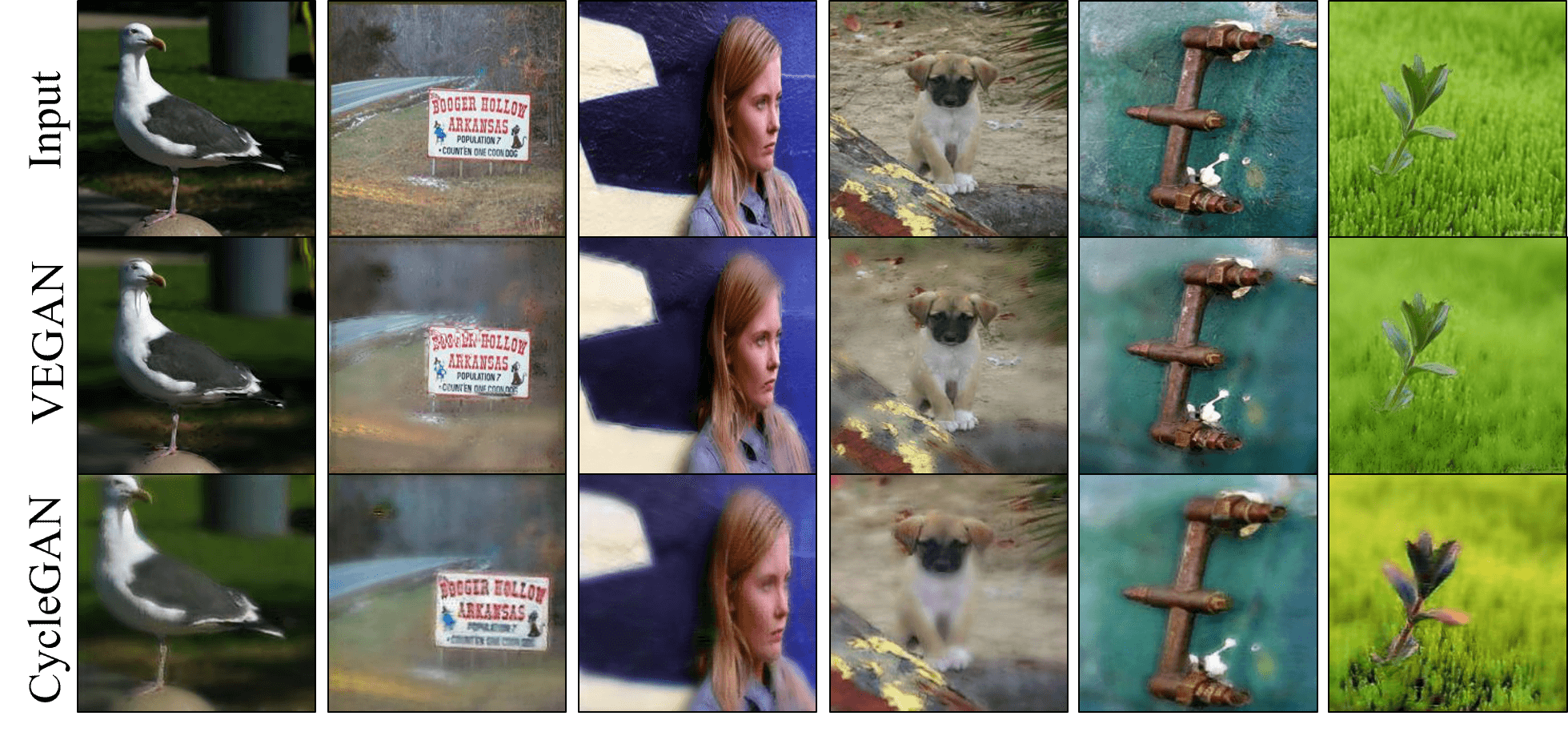} \\
    \vspace{3mm}
    `Defocus/Bokeh' visual effect generated by VEGAN (MSRA-D4) and CycleGAN. \vspace{-3mm}  
    \end{tabular}
    \caption{\label{fig:vsCycleGAN_BBCSDB_MSRA} The edited MSRA500 images generated by VEGAN and CycleGAN with respect to different expected visual effects.}
\end{figure*}

\begin{figure*}[t]
    \centering
    \begin{tabular}{c}
       \vspace{-2mm}
    \includegraphics[width=0.8\textwidth]{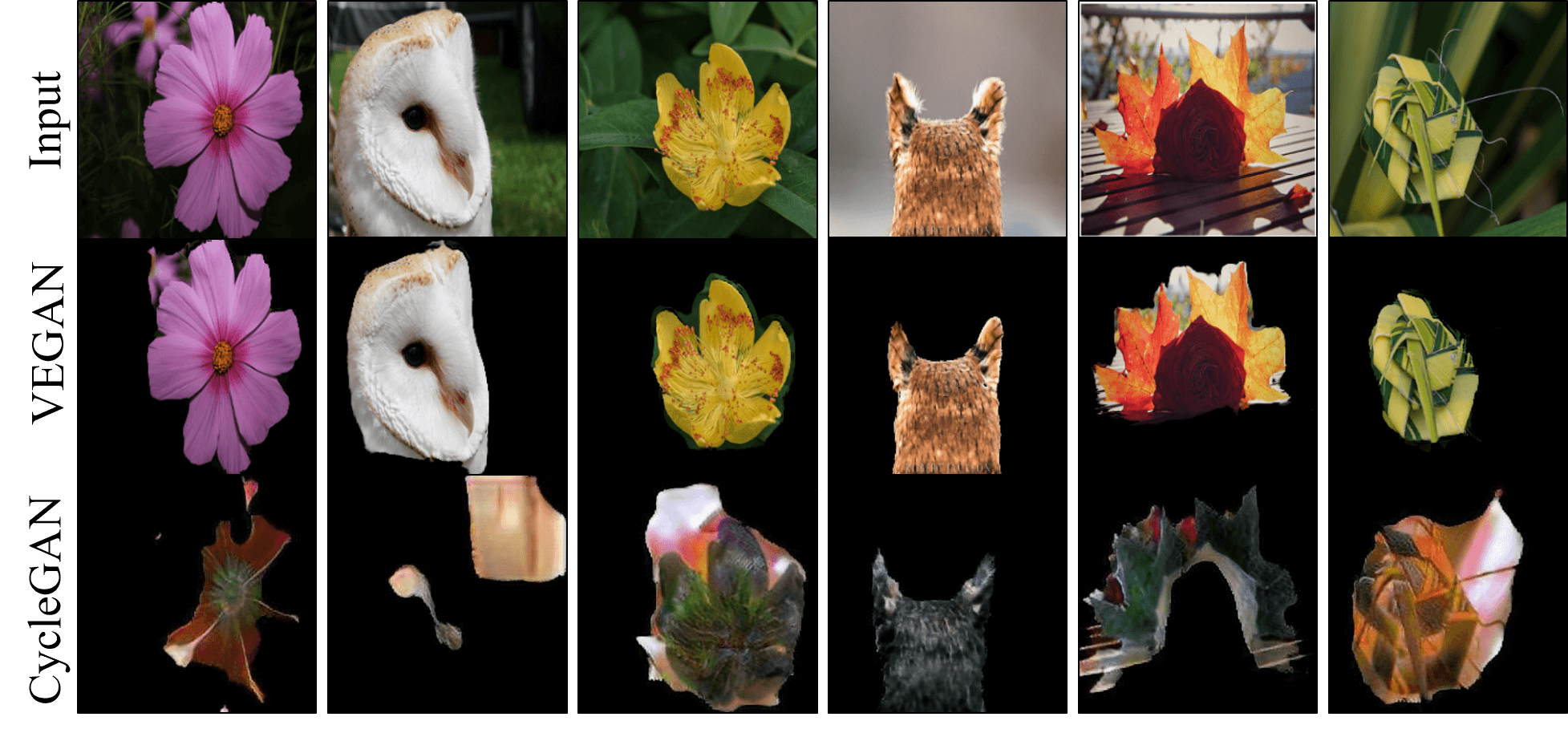} \\
       \vspace{3mm}
    `Black background' visual effect generated by VEGAN (Flickr-B4) and CycleGAN. \\
       \vspace{-2mm}
    \includegraphics[width=0.8\textwidth]{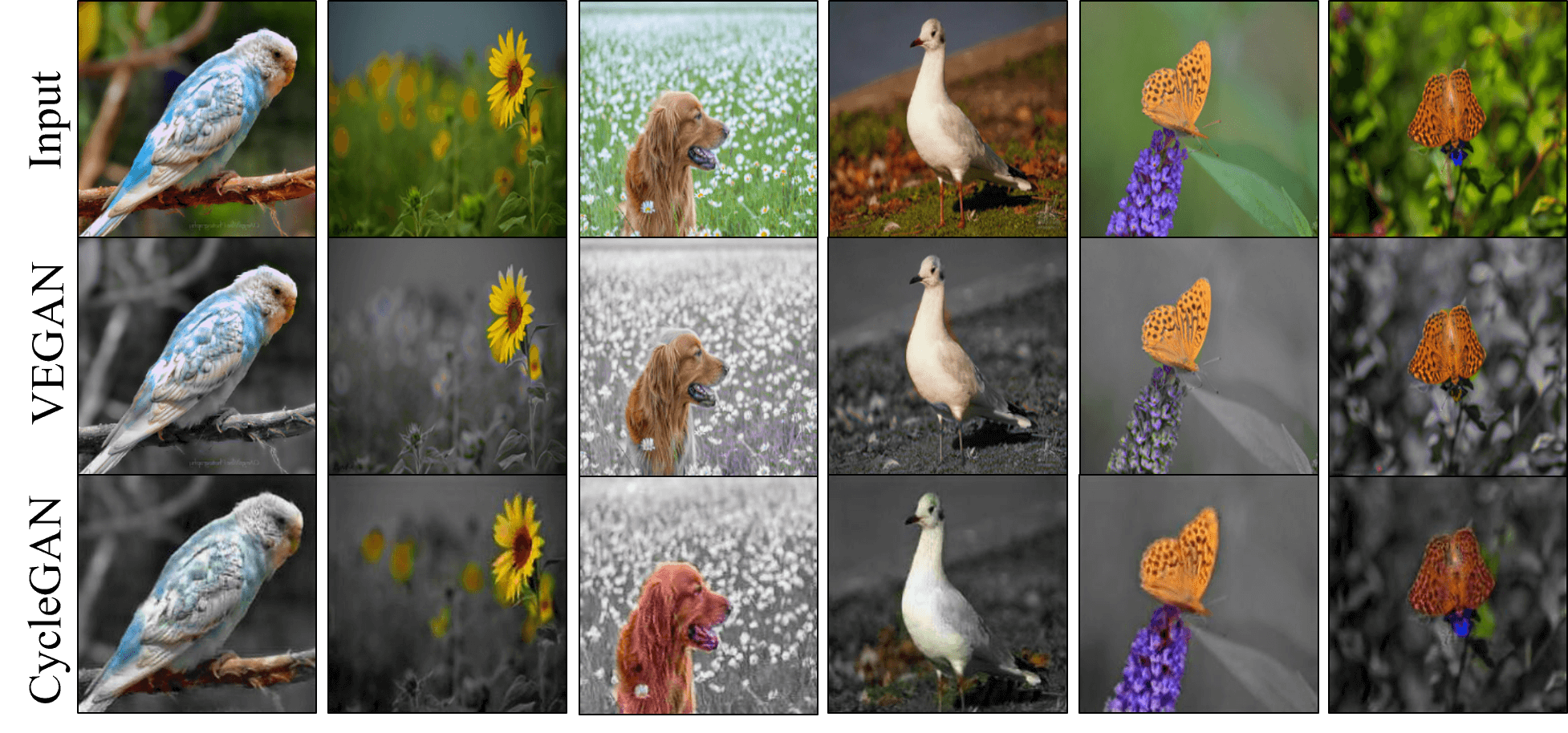} \\
       \vspace{3mm}
    `Color selectivo' visual effect generated by VEGAN (Flickr-C4) and CycleGAN.  \\   
       \vspace{-2mm}
    \includegraphics[width=0.8\textwidth]{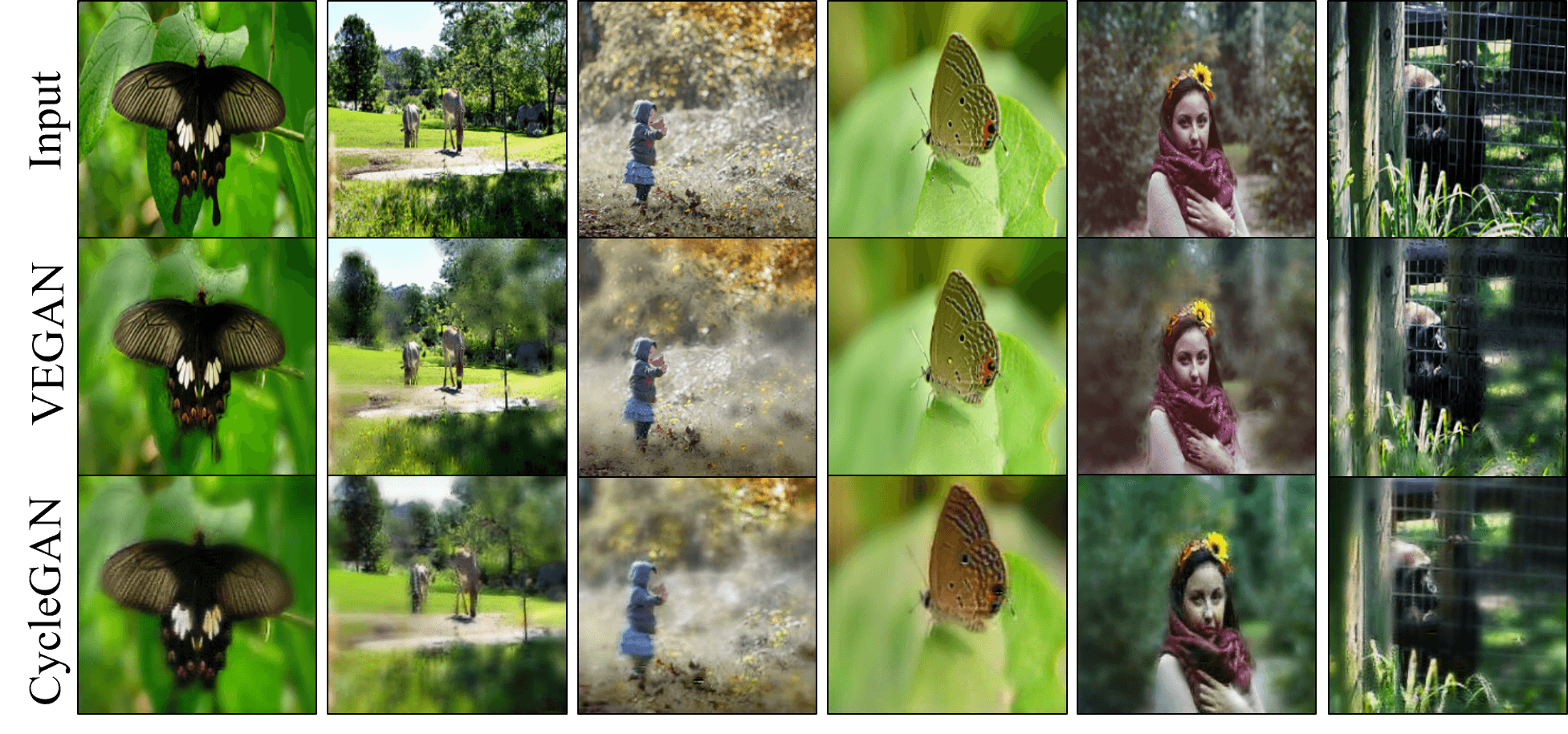} \\
       \vspace{3mm}
    `Defocus/Bokeh' visual effect generated by VEGAN (Flickr-D4) and CycleGAN. \vspace{-3mm}  
    \end{tabular}
    \caption{\label{fig:vsCycleGAN_BBCSDB_Flickr} The edited Flickr images generated by VEGAN and CycleGAN with respect to different expected visual effects.}
\end{figure*}

To train CycleGAN for visual-effect transfer, we use the set $\{I\}$ of original RGB images and the set $\{I_\mathrm{sample}\}$ of images with the expected visual effect as the two unpaired training sets. Fig.~\ref{fig:vsCycleGAN_BBCSDB_MSRA} shows the results of `training on MSRA9500 and testing on MSRA500'. Fig.~\ref{fig:vsCycleGAN_BBCSDB_Flickr} shows the results of `training on Flickr and testing on Flickr'. For CycleGAN and VEGAN, all the test images are unseen during training. The training process is done in an unsupervised manner without using any ground-truth annotations and paired images. 

Some comparison results are shown in Fig.~\ref{fig:vsCycleGAN_BBCSDB_MSRA} and Fig.~\ref{fig:vsCycleGAN_BBCSDB_Flickr}. We observe that the task of imitating {\tt black background} is actually more challenging for CycleGAN since the information of black regions in $\{I_\mathrm{sample}\}$ is limited and hence does not provide good inverse mapping back to $\{I\}$ under the setting of CycleGAN. 
The results of CycleGAN on imitating {\tt color selectivo} and {\tt defocus/Bokeh} are more comparable to those of VEGAN. However, the images generated by CycleGAN may have some distortions in color.  
On the other hand, VEGAN follows a well-organized procedure to learn how to imitate visual effects. The generator must produce a meaningful VER so that the editor can compose a plausible visual-effect image that does not contain noticeable artifacts for the discriminator to differentiate.

\end{document}